\documentclass{article}

\usepackage[final]{neurips_2025}
\usepackage{graphicx}

\usepackage[caption=false,font=footnotesize]{subfig}
\usepackage{wrapfig}

\usepackage{algorithm}
\usepackage{algorithmic}
\usepackage{booktabs}
\usepackage{multirow}
\usepackage{threeparttable}
\usepackage{makecell}
\usepackage{amsmath}
\usepackage{amsfonts}
\usepackage{pifont}
\usepackage{color}
\usepackage{colortbl}
\usepackage{newfloat}
\usepackage{listings}
\usepackage{subcaption}

\usepackage{xcolor}
\definecolor{S2Wcolor1}{HTML}{4e72b8}
\definecolor{W2Scolor1}{HTML}{f58220}
\colorlet{S2Wcolor}{S2Wcolor1!15}
\colorlet{W2Scolor}{W2Scolor1!10}

\usepackage[utf8]{inputenc} 
\usepackage[T1]{fontenc}    
\usepackage{hyperref}       
\usepackage{url}            
\usepackage{booktabs}       
\usepackage{amsfonts}       
\usepackage{nicefrac}       
\usepackage{microtype}      
\usepackage{xcolor}         

\title{Synergy Between the Strong and the Weak: \\Spiking Neural Networks are Inherently Self-Distillers}

\author{
  Yongqi Ding, Lin Zuo\thanks{Corresponding author.}, Mengmeng Jing, Kunshan Yang, Pei He, Tonglan Xie\\
  School of Information and Software Engineering\\
  University of Electronic Science and Technology of China\\
  \texttt{\{yqding,ksyang,202321090226,2021090907006\}@std.uestc.edu.cn}\\
  \texttt{linzuo@uestc.edu.cn, jingmeng1992@gmail.com}\\
}

\begin{document}

\maketitle

\begin{abstract}
Brain-inspired spiking neural networks (SNNs) promise to be a low-power alternative to computationally intensive artificial neural networks (ANNs), although performance gaps persist. Recent studies have improved the performance of SNNs through knowledge distillation, but rely on large teacher models or introduce additional training overhead. In this paper, we show that SNNs can be naturally deconstructed into multiple submodels for efficient self-distillation. We treat each timestep instance of the SNN as a submodel and evaluate its output confidence, thus efficiently identifying the strong and the weak. Based on this strong and weak relationship, we propose two efficient self-distillation schemes: (1) \textbf{Strong2Weak}: During training, the stronger "teacher" guides the weaker "student", effectively improving overall performance. (2) \textbf{Weak2Strong}: The weak serve as the "teacher", distilling the strong in reverse with underlying dark knowledge, again yielding significant performance gains. For both distillation schemes, we offer flexible implementations such as ensemble, simultaneous, and cascade distillation. Experiments show that our method effectively improves the discriminability and overall performance of the SNN, while its adversarial robustness is also enhanced, benefiting from the stability brought by self-distillation. This ingeniously exploits the temporal properties of SNNs and provides insight into how to efficiently train high-performance SNNs.
\end{abstract}

\section{Introduction}

Currently, artificial neural networks (ANNs) have shown outstanding performance in both computer vision and natural language processing~\cite{10854580,NEURIPS2023_204f828b}. However, ANN inference involves intensive multiply-accumulate (MAC) operations that consume significant power, limiting its deployment in power-constrained scenarios such as edge devices~\cite{SDT}. To reduce power consumption toward sustainable artificial intelligence, brain-inspired research is being conducted in anticipation of low-power artificial intelligence through neuromorphic computing~\cite{roy2019towards}.

Spiking Neural Networks (SNNs) are considered to be a low-power alternative to ANNs by transmitting information through bionic binary spikes, avoiding heavy MAC operations, and requiring only energy-efficient accumulation (AC) operations~\cite{SDT,SSNN}. Moreover, spiking neurons, which model the membrane potential dynamics of biological neurons that evolve over time, are able to extract underlying temporal features, making SNNs more favorable for exploiting their potential in temporal tasks~\cite{ponghiran2022spiking,kim2023exploring}. Combined with neuromorphic chips and dynamic vision sensors, SNNs can perform various tasks with ultra-low latency and power consumption~\cite{9543525}. For example, the SNN on the Tianmouc chip requires only 0.7mW to perform typical vision tasks~\cite{yang2024vision}.

However, limited by the binary information representation, there is still a performance gap between SNNs and ANNs. Further improving the performance of SNNs can facilitate their widespread deployment in real-world scenarios. Previous studies have optimized the spiking neurons~\cite{PLIF,GLIF,CLIF,PALIF}, network architectures~\cite{SDT,10535518,Shi_2024_CVPR}, spike-specific BN layers~\cite{TEBN,TAB}, and residual connections~\cite{fang2021deep,MSResNet,9597475}, significantly narrowing the performance gap. Another studies have focused on optimizing the training process of SNNs to be compatible with a variety of neurons and architectures, have received widespread attention~\cite{anumasa2024enhancing,hu2024highperformance,TRT}. Inspired by knowledge distillation, recent efforts have attempted to improve the performance of lightweight SNNs with additional teacher models (ANNs or SNNs)~\cite{KDSNN,9412147}. However, teacher models incur additional pre-training overhead and typically need to be customized for specific tasks, making the training process more cumbersome. To reduce the additional overhead, \cite{TSSD} extends the timestep during training and adds a weak classifier for spatio-temporal self-distillation. \cite{TKS} relies on real labels to guide incorrect output with correct output, which suffers from limited efficiency and performance. Given these unresolved issues, a question worth exploring is: \textit{How to distill a high-performance SNN without additional overhead?}

Following the philosophy of Occam's razor, we answer the above question by proposing a simple yet effective self-distillation strategy for SNNs. To begin with, we deconstruct the SNN in the temporal dimension, considering its instances at each timestep as a submodel. Then, we evaluate the output confidence of each submodel to identify the strong and the weak ones, respectively. Based on the strong and weak relationships, we propose two efficient self-distillation schemes: (1) \textbf{Strong2Weak} takes the strongest submodel as the "teacher" and guides the weakest "student" submodel to strengthen the discriminative ability of the weaker model, elevating the "short boards in the barrel" to improve the overall performance. (2) \textbf{Weak2Strong} works the other way around, where the weakest model serves as a "teacher'' to guide the strongest "student''. It is worth noting that Weak2Strong exploits the underlying dark knowledge of the weak "teacher" to augment the strong "student", e.g., an overly strong model may ignore detailed information leading to overfitting, while a weak model provides complements or acts as a regularizer, thus improving generalization~\cite{Yuan_2020_CVPR}. Both schemes allow flexible implementations, with default one-to-one distillation and alternative ensemble, simultaneous and cascade distillation. Moreover, these two distillation schemes are compatible with various network architectures and neurons with superior generalization. 

Visualizations and extensive experiments on both static and neuromorphic datasets show that the proposed self-distillation schemes narrow the performance gap between submodels of the SNN, improve overall performance, and enable inference with low latency. Meanwhile, the synergy between the strong and the weak promotes overall stability and enhances the robustness of the SNN against attacks. The contribution of this paper can be summarized as follows:
\begin{itemize}
    \item We show that SNNs can be deconstructed into multiple submodels in the temporal dimension, allowing for efficient self-distillation without any additional overhead.
    \item We efficiently identify the strong and the weak in the deconstructed submodels by confidence, and propose Strong2Weak and Weak2Strong self-distillation schemes to improve the performance of SNNs.
    \item Visualizations and experiments on both static and neuromorphic datasets show that the proposed self-distillation schemes enhance the discriminability, performance, and robustness of the SNN and enable satisfactory performance with ultra-low latency.
\end{itemize}

\section{Related work}

\textbf{Spiking Neural Network.} SNNs have received considerable attention due to their low-power characteristics~\cite{Su_2023_ICCV,bal2024spikingbert,Shen_2024_CVPR}, especially in light of the significant energy requirements associated with large language models. To improve the performance of SNNs, a plethora of studies have been conducted on spike encoding~\cite{qiu2024gated}, neuron models~\cite{PLIF,CLIF}, network architectures~\cite{SDT,10535518,Shi_2024_CVPR}, and training methods~\cite{anumasa2024enhancing,hu2024highperformance,TRT}. The contributions of this paper belong to training methods, compatible with various encoding, neuron models and network architectures with great generalization. Compared to other training methods, this paper deconstructs an SNN into multiple submodels and identifies the strong and weak submodels for self-distillation learning, which significantly improves performance without additional overhead.

\textbf{Knowledge Distillation.} Knowledge distillation originated in ANNs, which attempt to boost the performance of a weak student model with a strong teacher model~\cite{hinton2015distilling}, leaving only the student for inference without compromising inference efficiency. Depending on the spatial location of the distillation signal, distillation can be categorized into logit distillation~\cite{MLD,li2022curriculum} and feature distillation~\cite{romero2015fitnets,zong2023better}. Logit distillation encourages the student to mimic the teacher's output, and feature distillation promotes intermediate feature consistency between the two. Feature distillation typically offers superior performance, but the difficulty in obtaining the intermediate features of the teacher model in many scenarios (for privacy and security reasons) makes logit distillation more versatile and easier to implement~\cite{yuan2023studentfriendly,yang2023knowledge,Sun_2024_CVPR}. To improve distillation efficiency, self-distillation methods separate the teacher and student from a single model, e.g., by using intermediate checkpoints or adding multiple output heads~\cite{Yuan_2020_CVPR,BYOT}. This paper focuses on self-distillation learning in SNNs and shows that SNNs can naturally distinguish between the strong and the weak without any checkpoints or additional output heads, and are well suited for efficient self-distillation.

\textbf{Distillation-Enhanced SNN.} Given the remarkable gains of knowledge distillation in ANNs, the community has introduced it to train SNNs. Previous studies have shown that large ANNs and SNN teachers can improve the performance of small student SNNs~\cite{KDSNN,jointasnn,9412147,BKDSNN,guo2024enofsnn}.
\begin{wrapfigure}{r}[0cm]{0pt}        
    \includegraphics[width=0.55\linewidth]{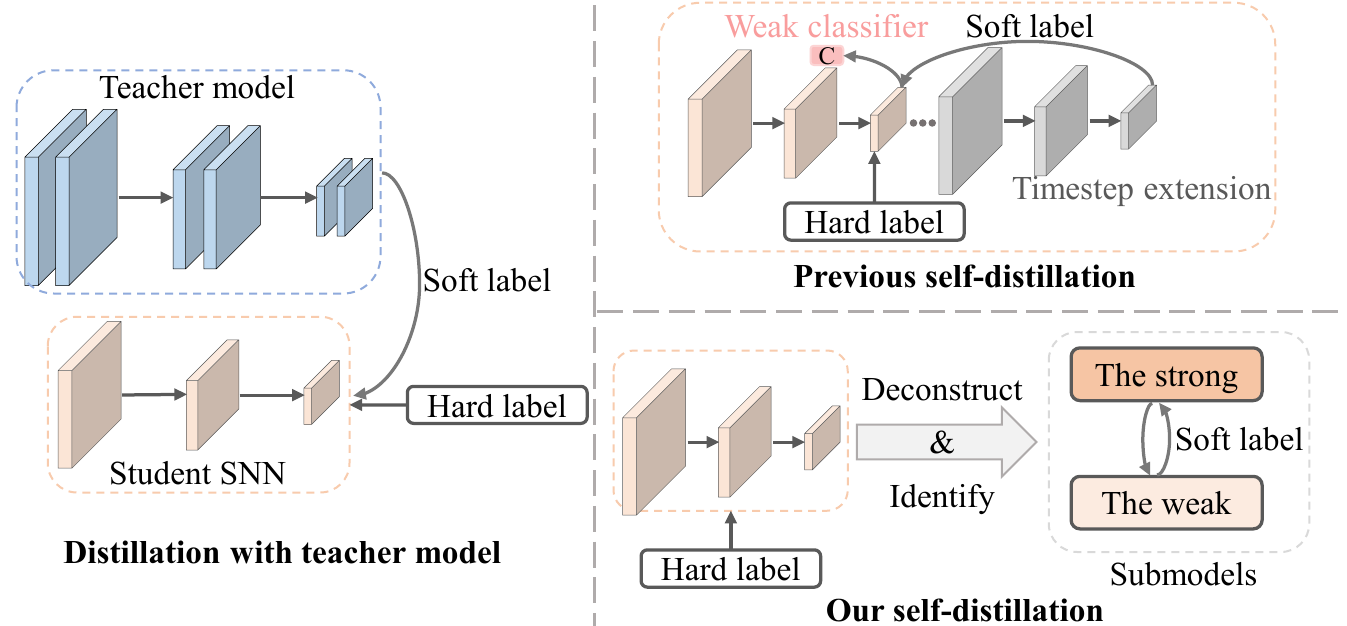}
    \caption{Comparison to other distillation methods. Our self-distillation deconstructs an SNN into multiple submodels and identifies the strong and weak ones for self-distillation without any additional overhead.} 
    \label{compare}
\end{wrapfigure}
However, the teacher model requires additional storage and computational overhead. To enhance the efficiency, self-distillation in SNNs has been explored. \cite{TKS} determines the correctness of the SNN output at each timestep, with the correct one guiding the incorrect one. However, this necessitates label information and cannot exploit the underlying dark knowledge in the error output, resulting in limited performance. \cite{TSSD} extends the training timestep and adds a additional weak classifier for spatio-temporal self-distillation, which improves the distillation performance but imposes a heavier training overhead. Unlike existing methods, our self-distillation schemes require no additional overhead and exploit both strong and weak dark knowledge with superior performance. A visual comparison with other distillation methods is shown in Fig.~\ref{compare}.

\section{Deconstructing the SNN}

\subsection{Temporal Properties of SNNs}

The core of the SNN lies in the spiking neurons, which endows it with low power consumption and temporal properties. Similar to biological neurons, spiking neurons maintain a membrane potential state, and the membrane potential changes continuously over time, depending on the input current received. In this paper, we use the most commonly used leaky integrate-and-fire (LIF)~\cite{STBP} neurons, but it is worth noting that our method can be applied to other neurons as well, since both have similar dynamics.

Let $H$ denote the membrane potential of the LIF neuron and $I$ indicate the input current generated by the previous layer of spikes and synaptic weights, the membrane potential dynamics of the LIF neuron can be expressed as:
\begin{equation}
H_{i}^{l}(t)=\left(1-\frac{1}{\tau}\right) H_{i}^{l}(t-1)+I_{i}^{l}(t),
\label{eq:charge}
\end{equation}
where $t$, $l$, and $i$ are the timestep, layer, and neuron indices, respectively. $\tau$ is the membrane potential time constant, which controls the degree of leakage of the membrane potential with time, i.e. the membrane potential of the LIF gradually decreases when there is no input.

If the membrane potential reaches the firing threshold $\vartheta$, the spiking neuron generates a spike with a value of 1, otherwise the output value is 0 (i.e. no spike is generated). After the spike is fired, the membrane potential is reset. This paper adopts the soft reset mechanism~\cite{Han_2020_CVPR}, i.e., the membrane potential is reduced by the same amount as the threshold:
\begin{equation}
H_{i}^{l}(t) = H_{i}^{l}(t)-S_{i}^{l}(t)\vartheta,
\end{equation}
where $S$ is the binary spike output.The residual membrane potential after reset remains until the next timestep to continue the dynamics (Eq.~\ref{eq:charge}), resulting in the initial membrane potential typically varying across timesteps. This residual membrane potential is considered to carry time-dependent information, so that the SNN can be used for temporal tasks~\cite{ponghiran2022spiking,kim2023exploring}.

\subsection{Deconstructing SNNs from the Temporal Dimension}

The output of a spike neuron at a given timestep depends not only on the initial membrane potential $H_{i}^{l}(t)$ but also on the received input $I_{i}^{l}(t)$. The difference between the two causes the output of the SNN to vary over timestep. Existing strategies average the outputs of multiple timesteps to improve overall output stability~\cite{STBP,SSNN}. However, we exploit this variability across timesteps to deconstruct an SNN into multiple submodels along the time dimension.

\begin{figure}[t]
\centering
\includegraphics[width=0.86\textwidth]{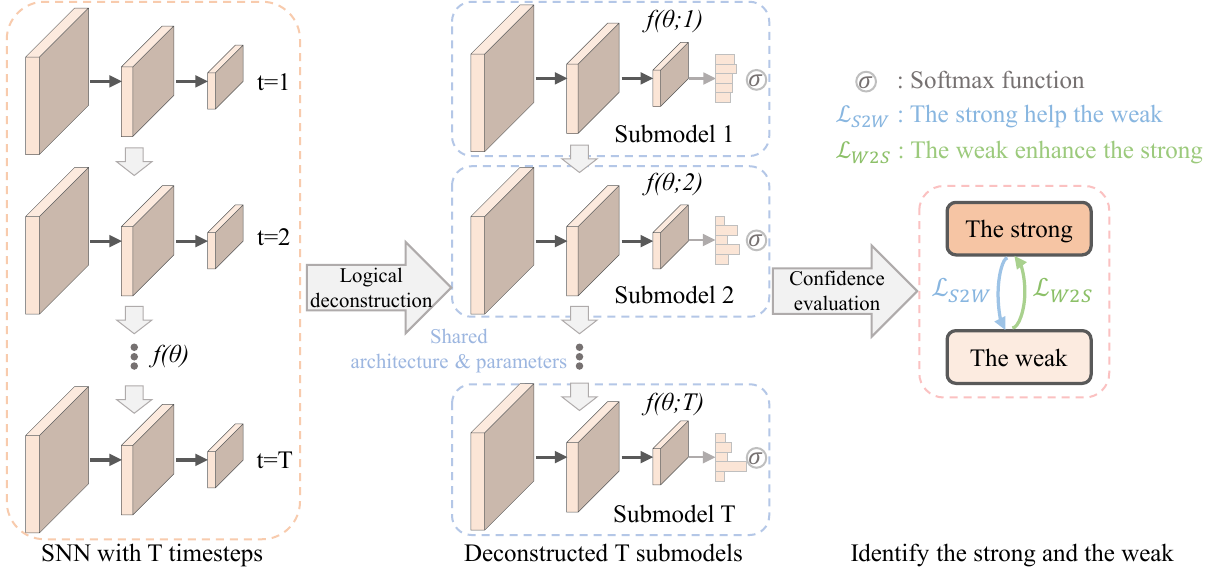}
\caption{Our method logically deconstructs the SNN with T timesteps into T submodels with the same architecture and parameters, and evaluates the confidence level of each submodel to identify the strong and the weak. The strong help the weak through distillation, and the weak transfer underlying dark knowledge to the strong, thus improving overall performance.}
\label{fig:overview}
\vspace{-0.5cm}
\end{figure}

Specifically, we consider the instances of the SNN at each timestep as a submodel. These submodels, while sharing the same architecture and parameters, can produce different outputs due to differences in neuron states and input currents. Let the SNN be $f(\theta)$, where $f(\cdot)$ and $\theta$ denote the architecture and parameters, respectively. Assuming that the SNN runs for a total of $T$ timesteps, we deconstruct it into $T$ submodels: $\{f(\theta;1),f(\theta;2),\cdots,f(\theta;T)\}$, as shown in Fig.~\ref{fig:overview}.

In this way, we deconstruct an SNN into multiple submodels with different outputs. These multiple distinct outputs form the necessary conditions for distillation learning, so that self-distillation can be performed without the need for other modules or strategies to generate additional outputs. Notably, this deconstruction is not limited to network architecture and neuron type, and is highly generalizable. In the following we describe how to perform self-distillation learning with the deconstructed submodels.

\section{Self-Distillation with Deconstructed SNNs}
\begin{wraptable}{r}{6.8cm}
\vskip -0.1in
\tabcolsep=0.005\columnwidth
  \caption{Accuracy comparison (\%) of different implementations.}
  \label{com_highacc1}
  \vskip -0.05in
  \scalebox{0.84}{
\begin{tabular}{ccc}
  \toprule
  Method  & CIFAR10-DVS & DVS-Gesture  \\
  \midrule
  High-accuracy teacher &  $78.43\pm0.33$ & $90.62\pm0.49$\\
  Low-loss teacher &  $78.93\pm0.33$ & $90.39\pm0.71$\\
  \hline
  \rowcolor{S2Wcolor}Strong2Weak (Ours) &  $78.93\pm0.12$ & $\textbf{91.43}\pm0.43$\\
  \rowcolor{W2Scolor}Weak2Strong (Ours) &  $\textbf{79.33}\pm0.29$ & $91.20\pm0.33$\\
  \bottomrule
 \end{tabular}
  }
\end{wraptable}
With the deconstructed multiple submodels, the key to distillation is to identify the teacher and student models. A naive distillation scheme is based on accuracy/loss, with a high-accuracy/low-loss submodel as the teacher guiding the low-accuracy/high-loss student. However, this manner (i) requires step-by-step comparisons with labels, which complicates the distillation process; (ii) cannot be used in scenarios where labels are missing, such as streaming online learning; and (iii) cannot exploit the underlying dark knowledge, resulting in limited performance, as shown in Table~\ref{com_highacc1}. To this end, we propose to quantify the output confidence of each submodel to efficiently evaluate the strong and weak relationships, and propose two distillation schemes based on the strong and weak relationships.

\subsection{Identify the Strong and the Weak}

Formally, let the output of submodel $f(\theta;t)$ be $o(t)=\{o_1(t),o_2(t),\cdots,o_C(t)\}$, where $C$ is the number of the object categories. We calculate its predicted probability distribution across the $C$ categories as
\begin{equation}
p(t)=\text{softmax}(o(t)).
\end{equation}
Then, we define the output confidence of the submodel as the maximum probability in $p(t)$:
\begin{equation}
con(t)=\text{max}(p(t)).
\end{equation}
As such, $con(t)$ reflects the degree of certainty of the submodel with respect to the output, similarly to~\cite{SEENN}.

Based on the output confidence $con(t)$, we evaluate the strong and weak relationships of these submodels. For any two submodels, we consider the one with the higher confidence to be the stronger one and the other to be the weaker one. Therefore, we can identify the strongest and the weakest among the $T$ submodels and thus exploit this strong and weak relationship for distillation learning. In particular, relying on confidence to evaluate strong and weak relationships does not require label information and is computationally simple without sacrificing efficiency.

\subsection{The Strong Help the Weak}

In classical knowledge distillation, the teacher model typically carries more parameters than the student and therefore better performance, i.e. strong teachers guide weak students~\cite{hinton2015distilling,MLD}. Accordingly, we propose Strong2Weak distillation: the strong submodel leads and enhances the weak submodel, thus improving overall performance.

Specifically, in each iteration we evaluate the confidence of each submodel and identify the strong and weak submodels with the highest and lowest confidence. The strong submodel is considered as the teacher and the weak submodel as the student, and logit distillation learning is performed between the two.

Formally, let the timestep index of the strong submodel be $t_s$ and that of the weak submodel be $t_w$. The output logits of the two submodels are $o(t_s)$ and $o(t_w)$ respectively. We soften the logit to obtain the probability distribution:
\begin{equation}
p(t_s)_j\!=\!\frac{e^{o(t_s)_j/\alpha}}{\sum^C_{c=1}{e^{o(t_s)_c/\alpha}}},p(t_w)_j\!=\!\frac{e^{o(t_w)_j/\alpha}}{\sum^C_{c=1}{e^{o(t_w)_c/\alpha}}},
\label{eq:callogit}
\end{equation}
where $\alpha$ is the softening factor, set to 2. We then use KL divergence to transfer knowledge from the strong model to the weak model:
\begin{equation}
\mathcal{L}_{S2W}\!\!=\!\!\alpha^2\!KL(p(t_s)||p(t_w))\!\!=\!\!\alpha^2\!\sum^{C}_{j=1}{p(t_s)_jlog(\!\frac{p(t_s)_j}{p(t_w)_j}\!)}.
\label{eq:s2wkl}
\end{equation}
During training of the SNN, the distillation loss and the cross entropy loss synergistically optimize the parameters. The total loss is therefore:
\begin{equation}
\mathcal{L}_{total}=\mathcal{L}_{CE}(O,Y) + \lambda_{S2W} \mathcal{L}_{S2W},
\end{equation}
where $Y$ is the label information and $\lambda_{S2W}$ is the coefficient, which defaults to 1.

During training, we use the rectangular surrogate gradient for optimization, since the spike activity is non-differentiable~\cite{STBP}. Thus, the derivative of the spike activity with respect to the membrane potential can be calculated as:
\begin{equation}
\frac{\partial S_{i}^{l}(t)}{\partial H_{i}^{l}(t)} \approx \frac{\partial h(H_{i}^{l}(t), \vartheta)}{\partial H_{i}^{l}(t)} = \frac{1}{a} \text{sign} (|H_{i}^{l}(t)-\vartheta|<\frac{a}{2}),
\end{equation}
where $a$ is a hyperparameter that controls the shape of the gradient and is set to 1.0.

\subsection{The Weak Enhance the Strong}

In the real world, teachers are also improved as they teach their students. In addition, it has been shown that weak models are able to learn underlying dark knowledge, which can contribute to the performance of stronger models~\cite{Yuan_2020_CVPR}. For example, an overly strong model may ignore detailed information and suffer from overfitting, while a weak model may provide complements or act as a regularizer to facilitate generalization~\cite{Yuan_2020_CVPR}. To this end, we propose the Weak2Strong distillation scheme: the weak submodel acts as the teacher, guiding the strong student to fully utilize the potential dark knowledge to improve overall performance.

Similar to Strong2Weak distillation, in Weak2Strong distillation we first identify the timestep indices of the strong and weak submodels to obtain their output logits $o(t_s)$ and $o(t_w)$. Then we compute the two corresponding probability distributions according to Eq.~\ref{eq:callogit}. When calculating the distillation loss, we use $p(t_w)$ as the teacher to prompt $p(t_s)$ to learn the dark knowledge in $o(t_w)$:
\begin{equation}
\mathcal{L}_{W2S}\!\!=\!\!\alpha^2\!KL(p(t_w)||p(t_s))\!\!=\!\!\alpha^2\!\sum^{C}_{j=1}{p(t_w)_jlog(\frac{p(t_w)_j}{p(t_s)_j})}.
\label{eq:w2skl}
\end{equation}

Combined with the cross-entropy loss, the final loss is obtained as:
\begin{equation}
\mathcal{L}_{total}=\mathcal{L}_{CE}(O,Y) + \lambda_{W2S} \mathcal{L}_{W2S},
\end{equation}
where $\lambda_{W2S}$ is the hyperparameter coefficient set to 1.

The schematic diagrams of these two distillation schemes are shown in Fig.~\ref{fig:overview}. In this way, self-distillation can be achieved without additional teacher models or auxiliary modules, and we provide pseudocode for Strong2Weak and Weak2Strong distillation in Alg.~\ref{alg: strong2weak}. Note that in identifying the strong and weak submodels, we take the confidence mean of a batch of samples rather than distilling sample by sample.

\vspace{-0.1in}
\subsection{Flexible Self-Distillation Implementation}
The above describes the default implementation of the Strong2Weak (S2W) and Weak2Strong (W2S) self-distillation schemes, i.e., one-to-one distillation of the strongest and weakest submodels with KL divergence. In addition, we would like to emphasize that the deconstructed SNN allows flexible distillation implementations with great extensibility. Below we discuss two flexible components that can be reimplemented: distillation losses and teacher-student configurations.
student configurations.

\begin{wraptable}[8]{r}{7cm}
\vspace{-0.2in}
  \caption{Performance with flexible distillation loss functions (\%).}
  \label{com_loss}
  \tabcolsep=0.005\columnwidth
  \scalebox{0.84}{
\begin{tabular}{ccc|cc}
  \toprule
 \multirow{2}{*}{Loss function} & \multicolumn{2}{c}{CIFAR10-DVS} & \multicolumn{2}{c}{DVS-Gesture}\\ & S2W & W2S & S2W & W2S\\
  \midrule
  KL divergence & 78.93 & 79.33 & 91.43 & 91.20\\
  MSE & 76.50 & 76.67 & 92.47 & 91.44\\
  Logit standardization~\cite{Sun_2024_CVPR} & 75.93 & 75.77 & 91.67 & 91.55\\
  \bottomrule
 \end{tabular}}
\end{wraptable}

\colorbox{blue!5}{\textbf{Remark1}:\textit{ Flexible distillation loss function.}} In Eq.~\ref{eq:s2wkl} and Eq.~\ref{eq:w2skl} we use the most commonly used KL divergence as the distillation loss. However, as shown in Table~\ref{com_loss}, our method is able to seamlessly support other distillation functions (e.g., MSE and Logit standardization~\cite{Sun_2024_CVPR}) with consistently superior performance. This flexibility allows practitioners to further improve performance by incorporating specially designed loss functions~\cite{MLD,li2022curriculum,Sun_2024_CVPR} based on task requirements, establishing our method as a generalized substrate for high-performance SNNs.

\begin{wraptable}[10]{r}{6cm}
  \caption{Performance with flexible implementations (\%).}
  \label{com_config}
  \vspace{-0.08in}
  \tabcolsep=0.005\columnwidth
  \scalebox{0.84}{
\begin{tabular}{ccc|cc}
  \toprule
 \multirow{2}{*}{Implementation} & \multicolumn{2}{c}{CIFAR10-DVS} & \multicolumn{2}{c}{DVS-Gesture}\\ & S2W & W2S & S2W & W2S\\
  \midrule
  Default & 78.93 & 79.33 & 91.43 & 91.20\\
  Ensemble teacher & 78.53 & 79.17 & 92.02 & 91.78\\
  Ensemble student & 78.30 & 78.37 & 91.55 & 91.55\\
  Simultaneous & \multicolumn{2}{c}{79.17} & \multicolumn{2}{c}{91.32}\\
  Cascade & 79.03 & 79.37 & 93.41 & 92.13\\
  \bottomrule
 \end{tabular}
  }
\end{wraptable}

\colorbox{blue!5}{\textbf{Remark2}:\textit{ Flexible teacher and student configurations.}} At each iteration, we distill between the strongest and weakest submodels. Since the timestep is usually greater than 2, there are options to use more submodels to compose the teacher or student in the implementation. For example, an average of the $T-1$ higher/lower confidence submodels can be used as the teacher to guide the weakest/strongest, and we refer to this implementation as the ensemble teacher. Similarly, a highest/lowest confidence submodel can also guide the remaining $T-1$ submodels, i.e., the ensemble student. Alternatively, both S2W and W2S distillations can be performed simultaneously, or cascade distillation can be performed depending on the confidence level. We detail and analyze these alternative implementations in \textbf{Appendix~\ref{app_A}}. Table~\ref{com_config} shows the performance with various implementations, indicating that flexible implementations can further improve performance, especially for cascade distillation. It is worth noting that the simultaneous use of S2W and W2S did not achieve significantly better performance because the excessive similarity between the submodels reduces the overall diversity of the SNN, thus limiting generalizability. How to balance diversity and similarity between multiple submodels is a perennial problem in ensemble learning~\cite{9677845,NEURIPS2020_b86e8d03}, and we leave it for future work.

It is worth noting that while our proposed method of self-distillation after deconstruction offers wide scope for extension and implementation, the performance under different implementations is often affected by various factors such as task, data, and experimental setup. Therefore, we have kept the default configuration in the following experiments to highlight the superiority of our efficient distillation methods rather than the specific implementation, leaving further extensions for future work. More importantly, experimental results show that even the default implementation, which is not deliberately tuned, can significantly improve the performance of vanilla SNNs, especially for neuromorphic tasks with temporal dimensions.

\section{Experiments}
\label{exp}
We perform experiments on both static images CIFAR10/100~\cite{CIFAR}, ImageNet and neuromorphic datasets CIFAR10-DVS~\cite{CIFAR10-DVS}, DVS-Gesture~\cite{DVS-Gesture} to confirm the effectiveness and performance advantages of the proposed self-distillation schemes. Both VGG and MS-ResNet~\cite{MSResNet}, and Transformer architectures are used in the experiments. To reduce the influence of randomness, we report the average results of three trials. The detailed experimental setup is presented in \textbf{Appendix~\ref{app_B}}.

\subsection{Ablation Studies}
\label{ablationstudy}

\begin{wraptable}[17]{r}{6.9cm}
\vspace{-0.6cm}
\tabcolsep=0.001\columnwidth
  \caption{Ablation studies. The proposed self-distillation schemes show consistent performance gains, especially on neuromorphic datasets with temporal properties.}
  \label{tab:ablation}
  \begin{threeparttable}
  \scalebox{0.88}{
  \begin{tabular}{c|c|cc}
    \toprule
    \multirow{2}{*}{Dataset} & \multirow{2}{*}{Method} & \multicolumn{2}{c}{Accuracy (\%)} \\
    & & VGG-9 & MS-ResNet18\\ 
    \midrule
    \rowcolor{gray!8} \cellcolor{white}& Vanilla SNN & $94.21$ & $94.88$\\
    \rowcolor{S2Wcolor} \cellcolor{white}& Strong2Weak & $\textbf{94.79}_{\textbf{+0.58}}$  & $\textbf{95.15}_{\textbf{+0.27}}$\\
   \rowcolor{W2Scolor} \cellcolor{white}\multirow{-3}{*}{CIFAR10}& Weak2Strong & $94.70_{+0.49}$ & $95.13_{+0.25}$\\
    \hline
    \rowcolor{gray!8} \cellcolor{white}& Vanilla SNN & $74.41$ & $76.33$\\
    \rowcolor{S2Wcolor} \cellcolor{white}& Strong2Weak & $76.02_{+1.61}$  & $\textbf{78.25}_{\textbf{+1.92}}$ \\
    \rowcolor{W2Scolor} \cellcolor{white}\multirow{-3}{*}{CIFAR100}& Weak2Strong & $\textbf{76.16}_{\textbf{+1.75}}$ & $77.98_{+1.65}$\\
    \hline
   \rowcolor{gray!8} \cellcolor{white}& Vanilla SNN & $73.97$ & $66.40$\\
    \rowcolor{S2Wcolor} \cellcolor{white}& Strong2Weak & $78.93_{+4.96}$ & $70.50_{+4.10}$\\
    \rowcolor{W2Scolor} \cellcolor{white}\multirow{-3}{*}{CIFAR10-DVS}& Weak2Strong & $\textbf{79.33}_{\textbf{+5.36}}$ & $\textbf{71.57}_{\textbf{+5.17}}$\\
    \hline
    \rowcolor{gray!8} \cellcolor{white}& Vanilla SNN & $87.85$ & $89.35$\\
    \rowcolor{S2Wcolor} \cellcolor{white}& Strong2Weak & $\textbf{91.43}_{\textbf{+3.58}}$ & $90.86_{+1.51}$\\
    \rowcolor{W2Scolor} \cellcolor{white}\multirow{-3}{*}{DVS-Gesture}& Weak2Strong & $91.20_{+3.35}$ & $\textbf{91.55}_{\textbf{+2.20}}$\\
    \bottomrule
  \end{tabular}}
  \end{threeparttable}
\end{wraptable}

In Table~\ref{tab:ablation}, we compare Strong2Weak and Weak2Strong to the vanilla SNN on static and neuromorphic datasets. The results show that for both VGG and MS-ResNet architectures, Strong2Weak and Weak2Strong are able to provide consistent performance gains. In particular, recognition accuracy on CIFAR10-DVS was improved by up to 5.26\%, a more significant improvement than on static datasets. This is due to the lack of temporal features in static images, so the performance differences between the deconstructed submodels are minimal and do not maximize the distillation effect. Neuromorphic data, on the other hand, is rich in temporal features, taking full advantage of the mutually enhancing effect of strong and weak submodels. This sheds light on the potential of our method to be applied to more temporal tasks in the future.

\begin{figure}[t]
\centering
\subfloat[Vanilla SNN]
	{
	\includegraphics[width=0.2\textwidth]{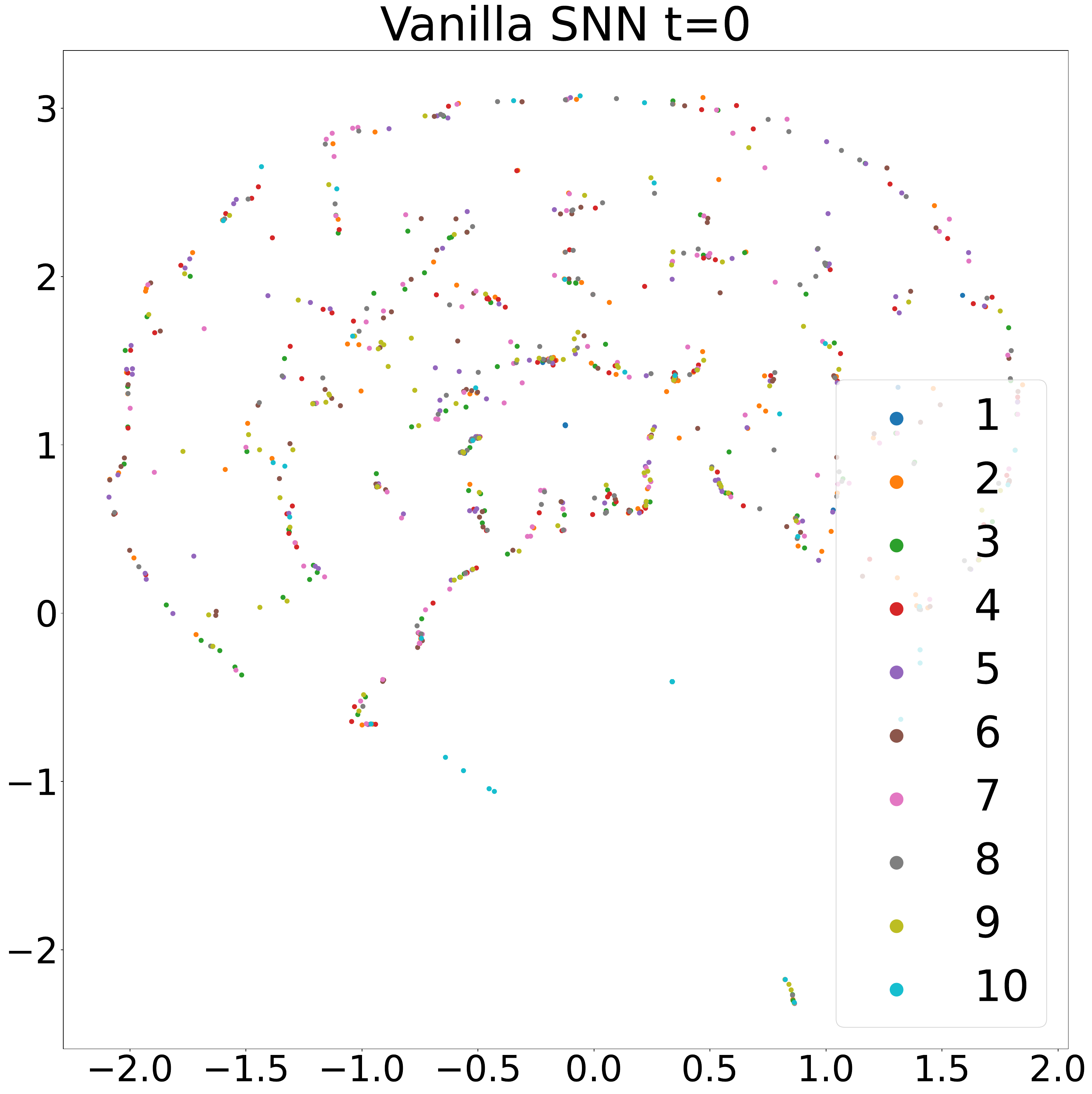}\hspace{-1mm}
	\includegraphics[width=0.2\textwidth]{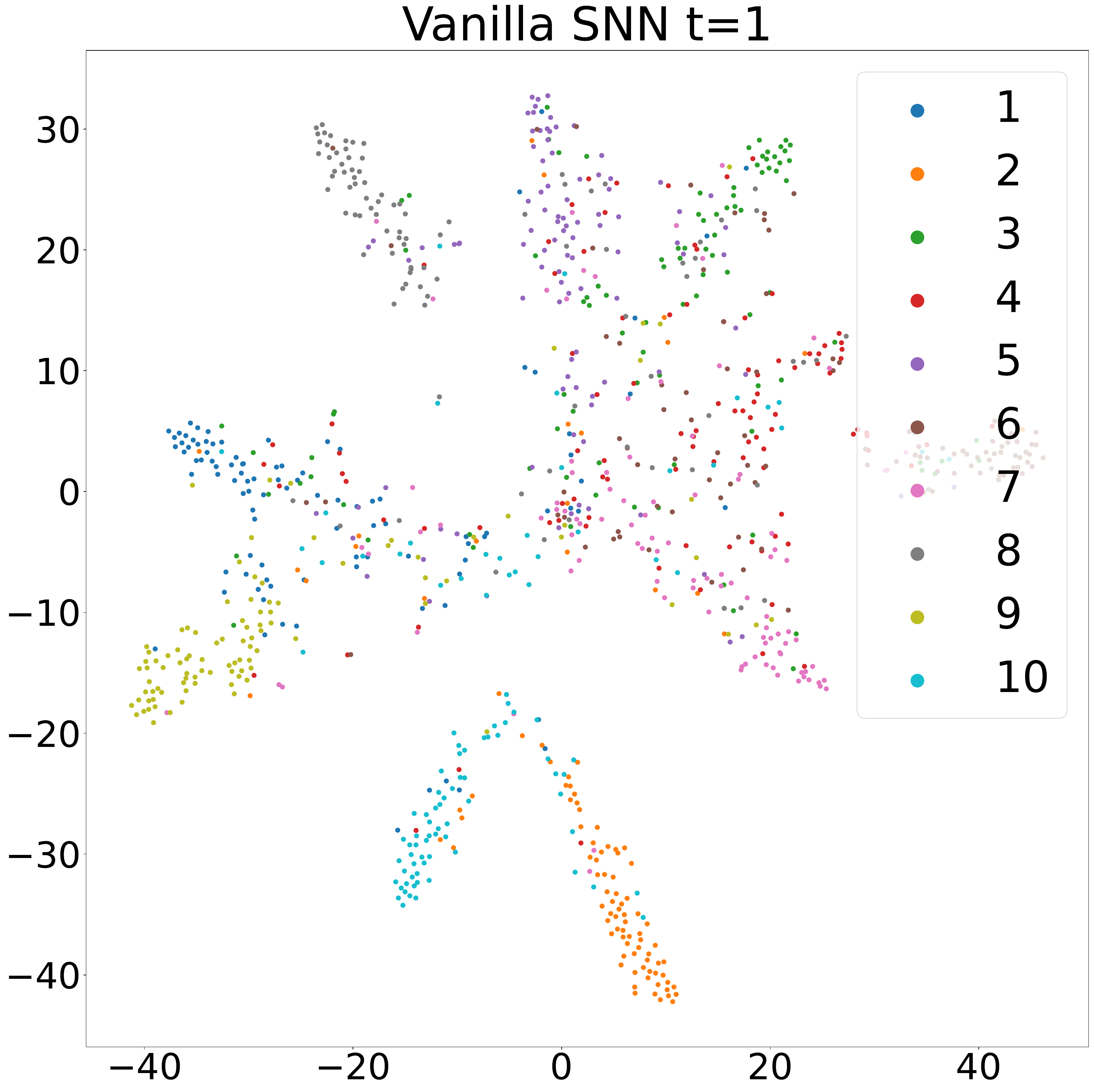}\hspace{-1mm}
	\includegraphics[width=0.2\textwidth]{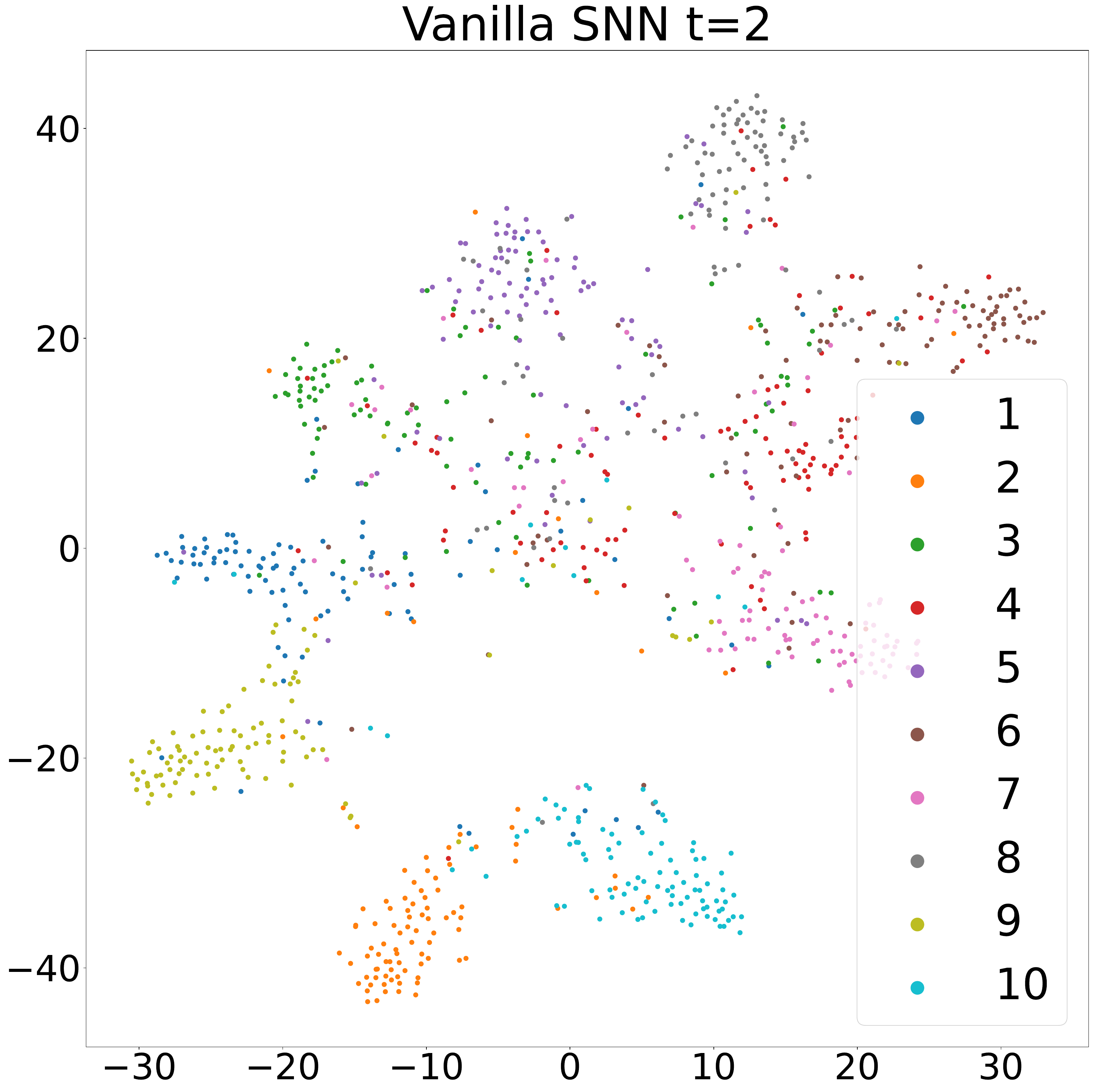}\hspace{-1mm}
	\includegraphics[width=0.2\textwidth]{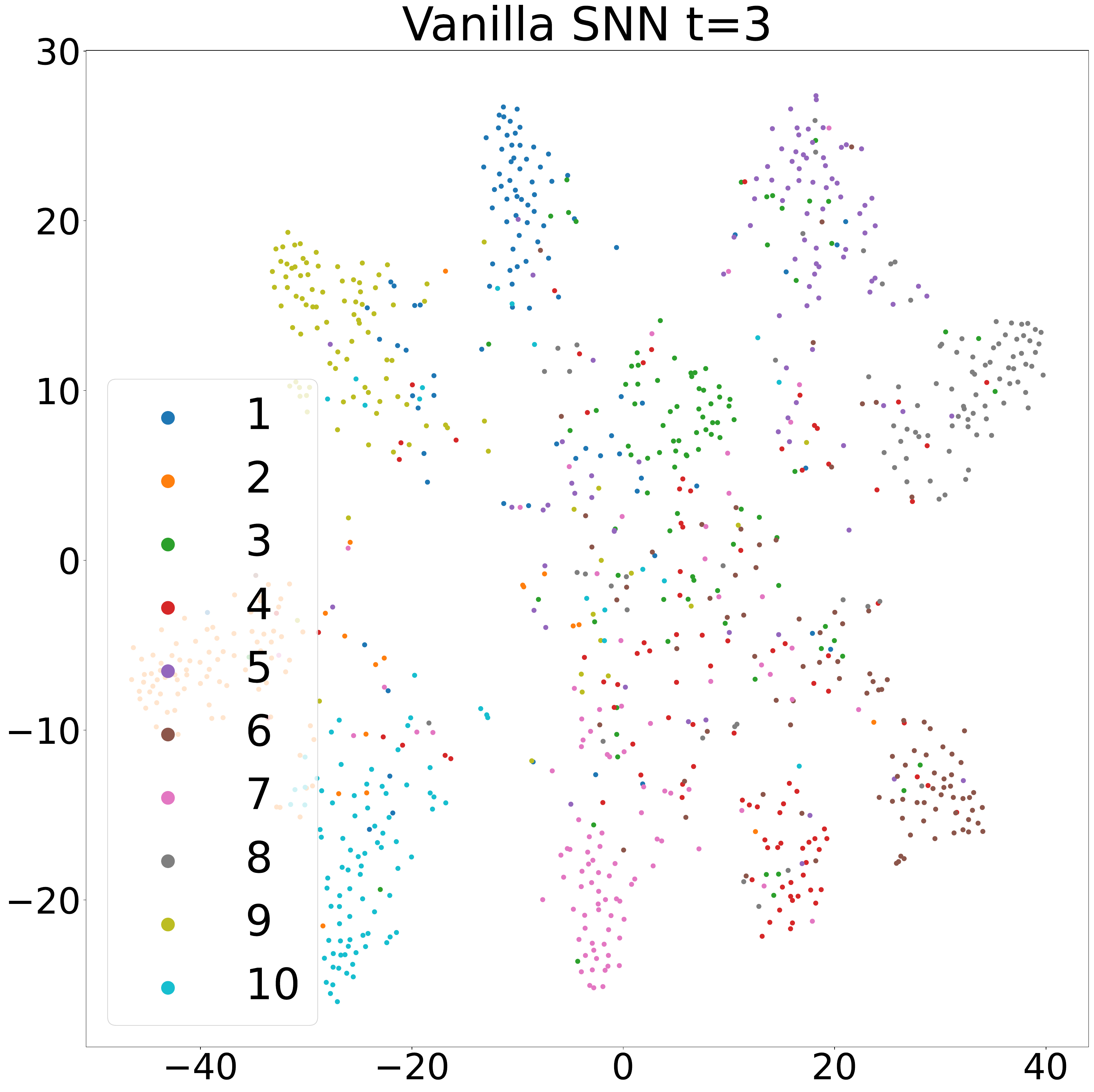}\hspace{-1mm}
	\includegraphics[width=0.2\textwidth]{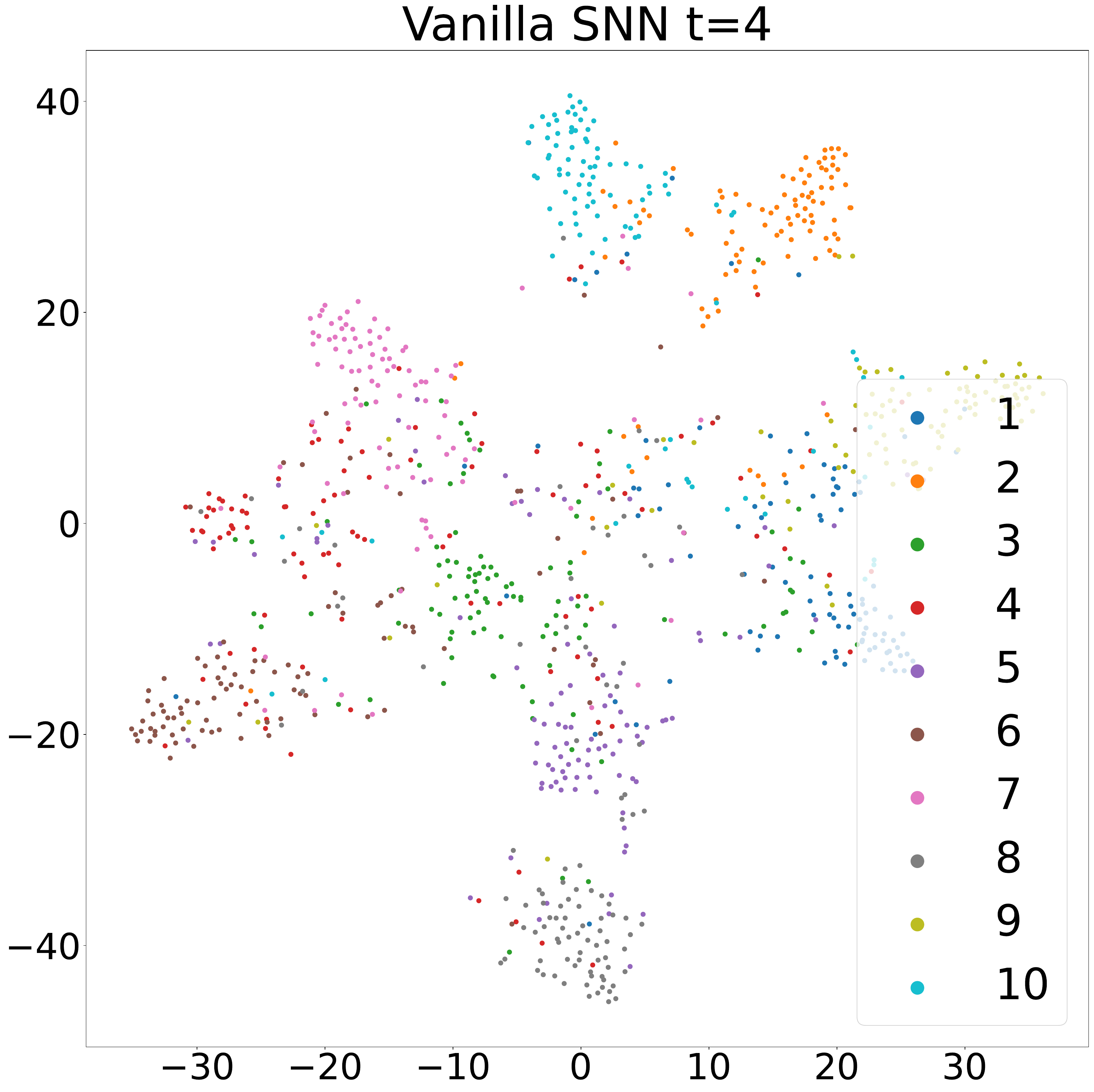}
	}
\vspace{-0.15in}
\subfloat[Strong2Weak]
	{
	\includegraphics[width=0.2\textwidth]{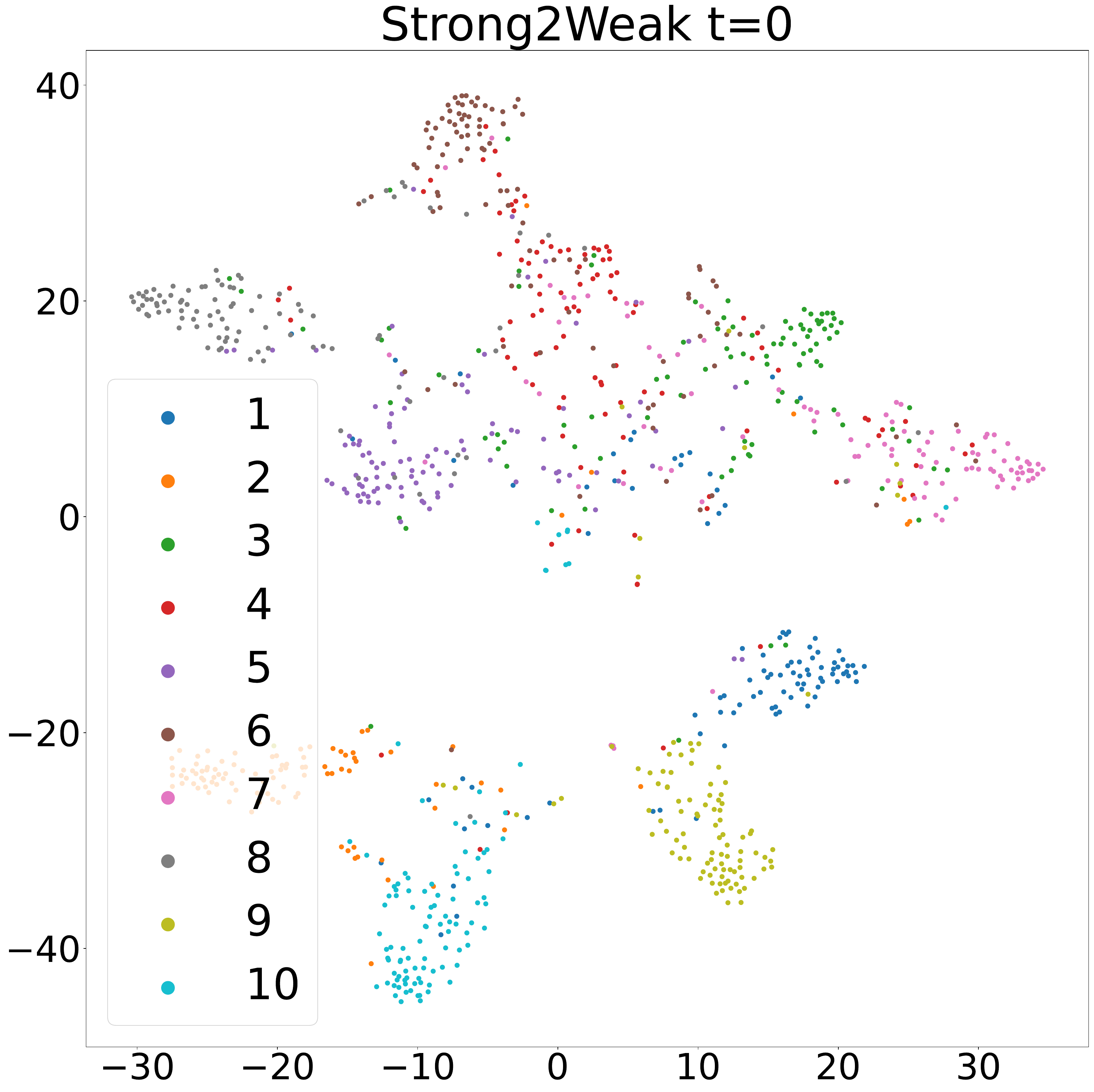}\hspace{-1mm}
	\includegraphics[width=0.2\textwidth]{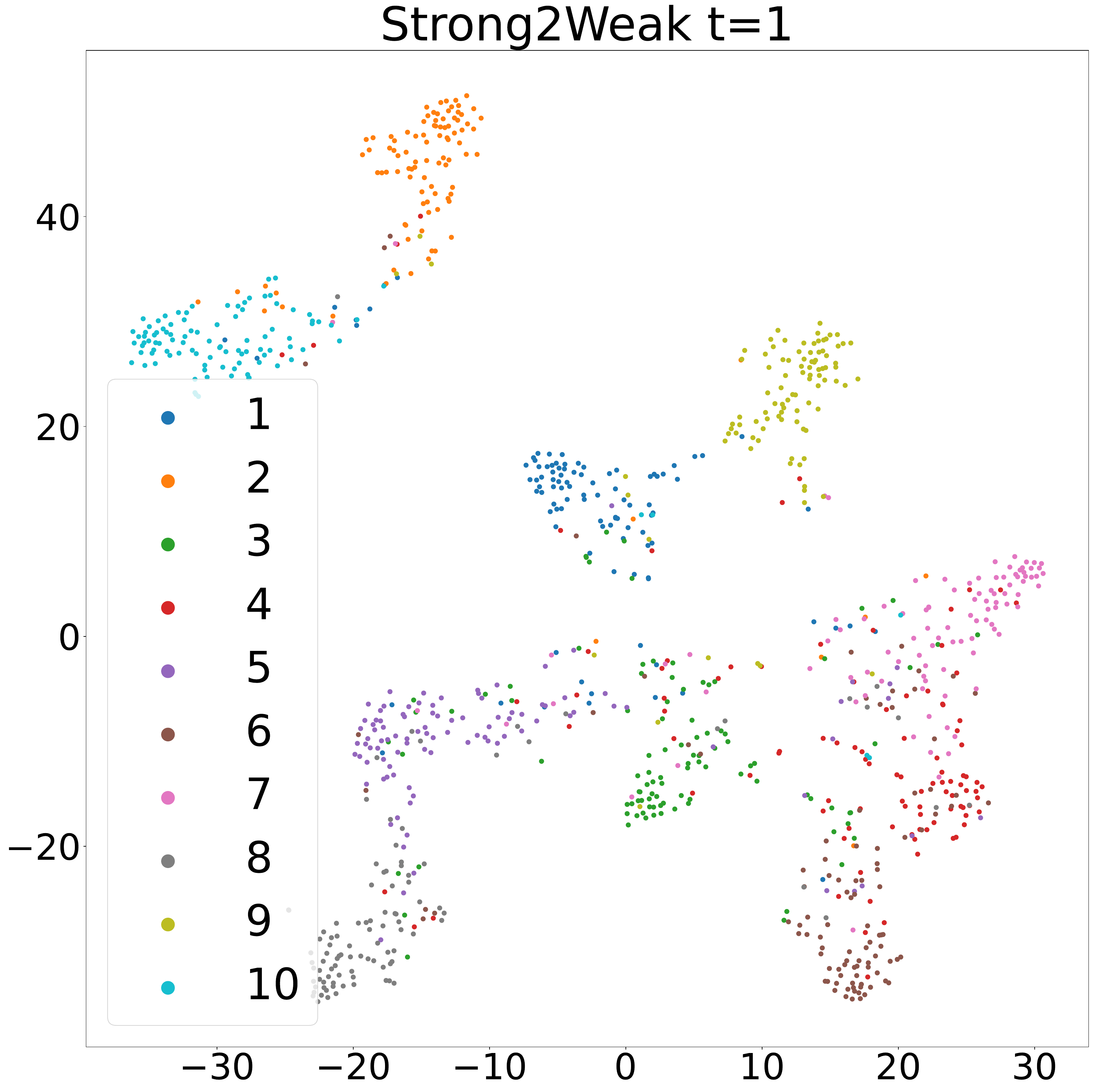}\hspace{-1mm}
	\includegraphics[width=0.2\textwidth]{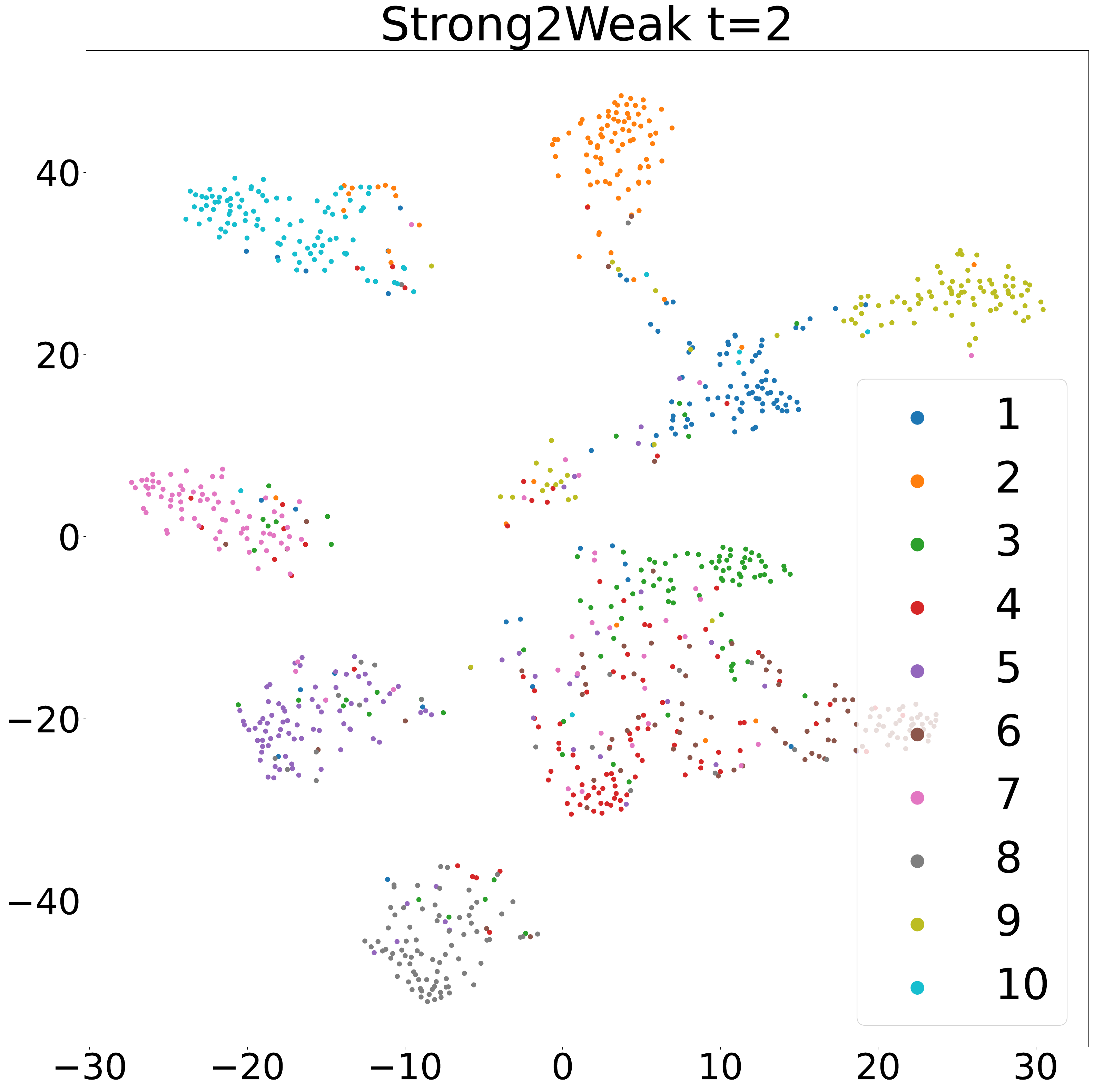}\hspace{-1mm}
	\includegraphics[width=0.2\textwidth]{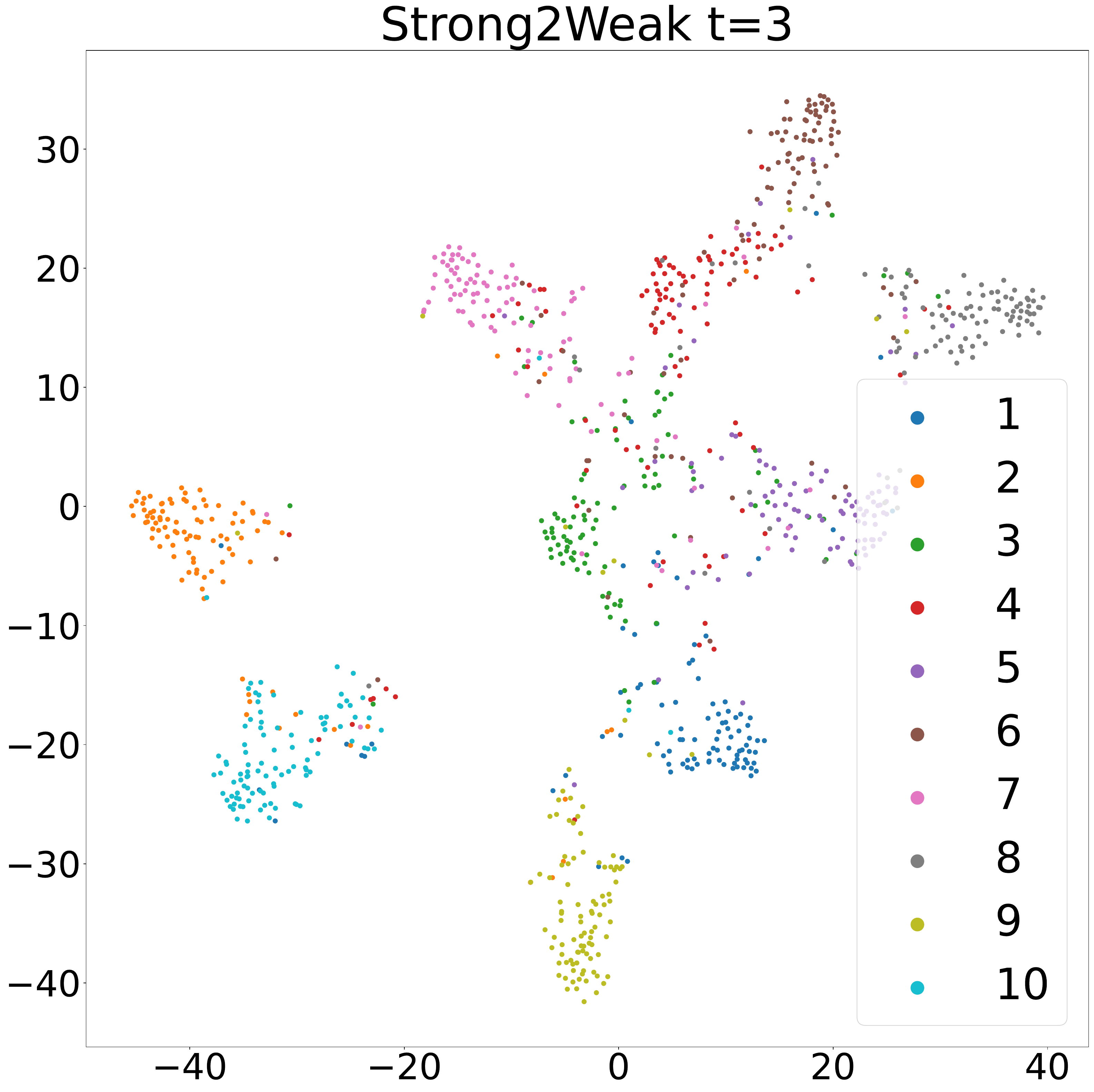}\hspace{-1mm}
	\includegraphics[width=0.2\textwidth]{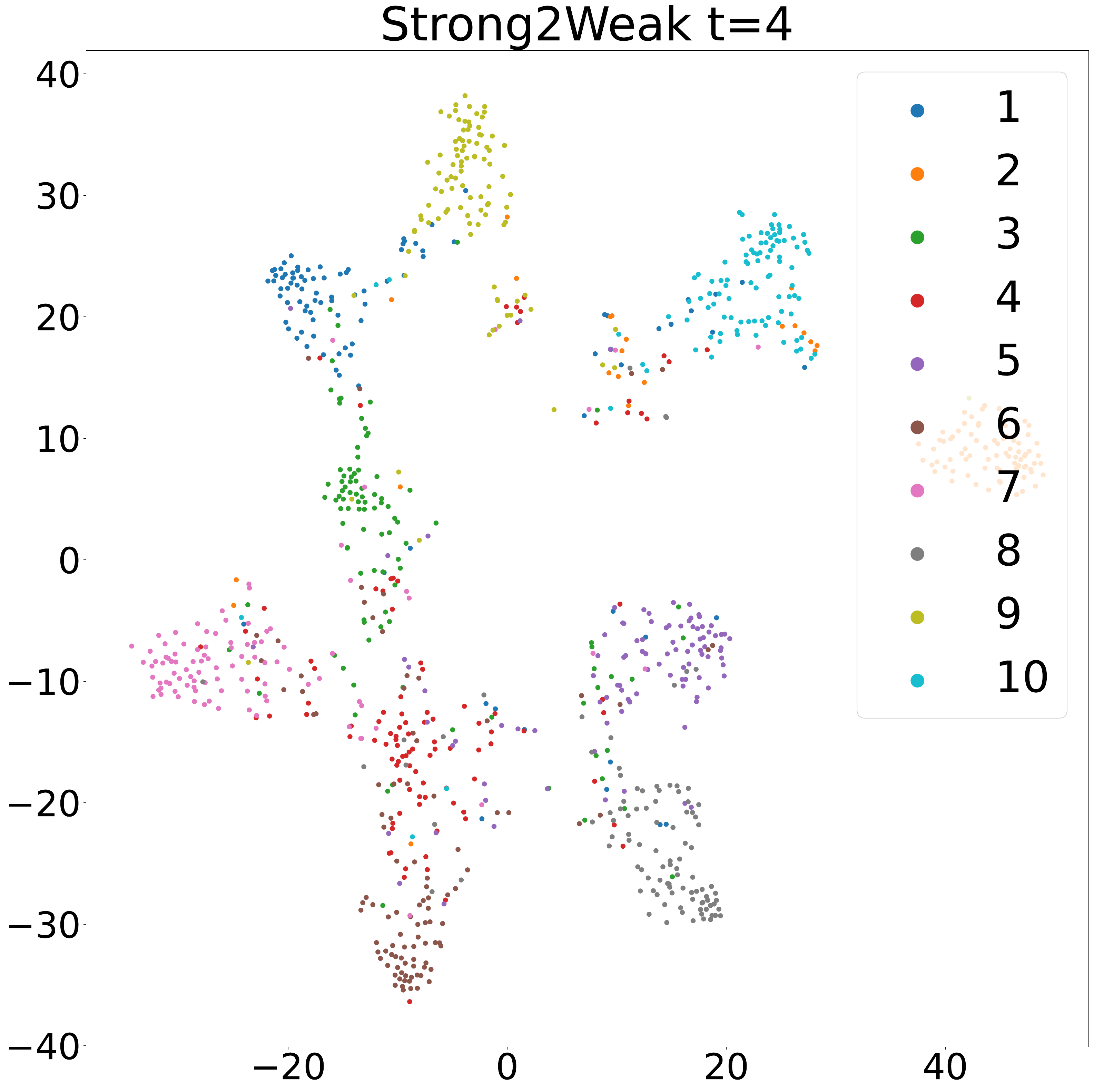}
	}
\vspace{-0.15in}
\subfloat[Weak2Strong]
	{
	\includegraphics[width=0.2\textwidth]{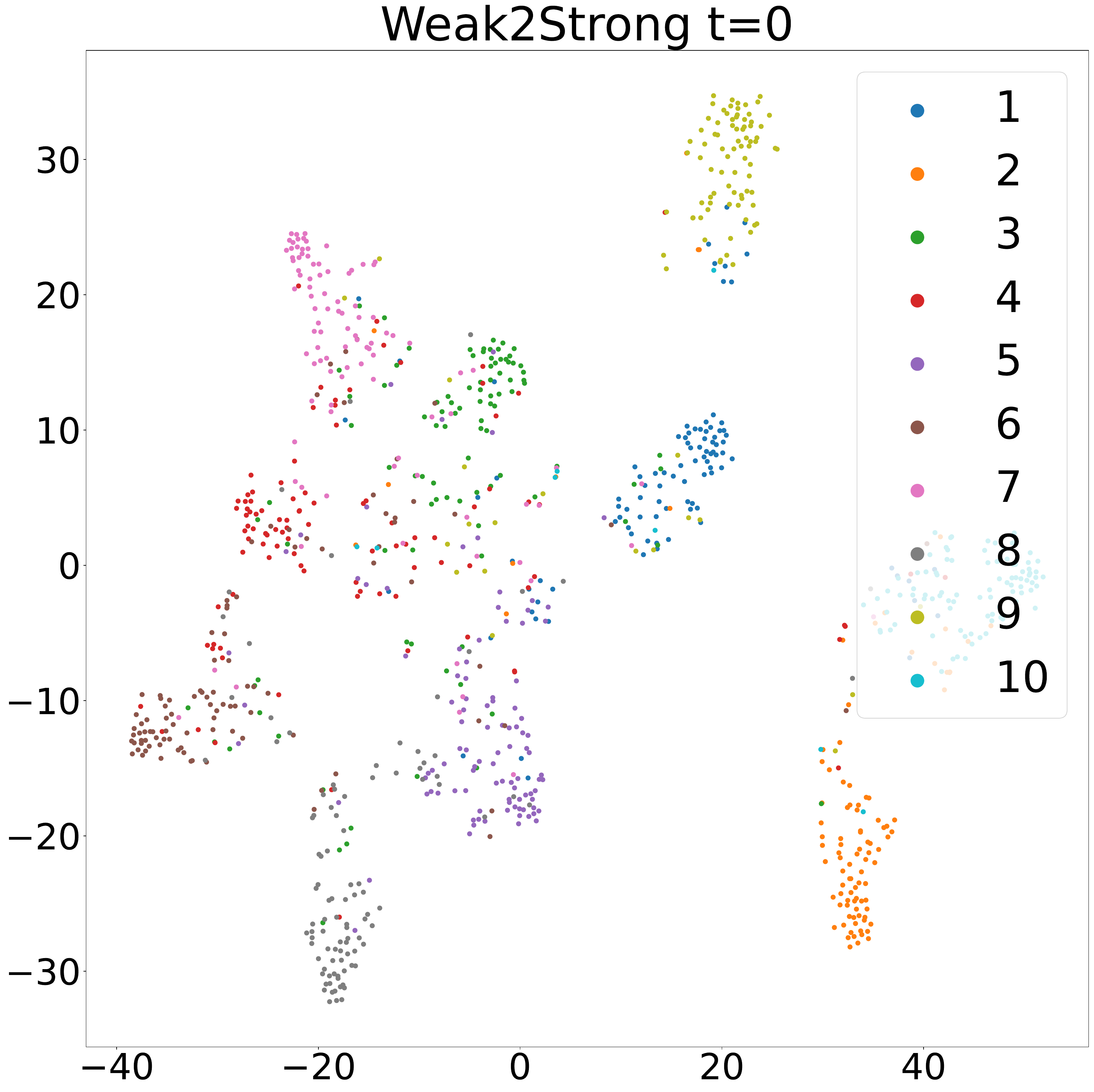}\hspace{-1mm}
	\includegraphics[width=0.2\textwidth]{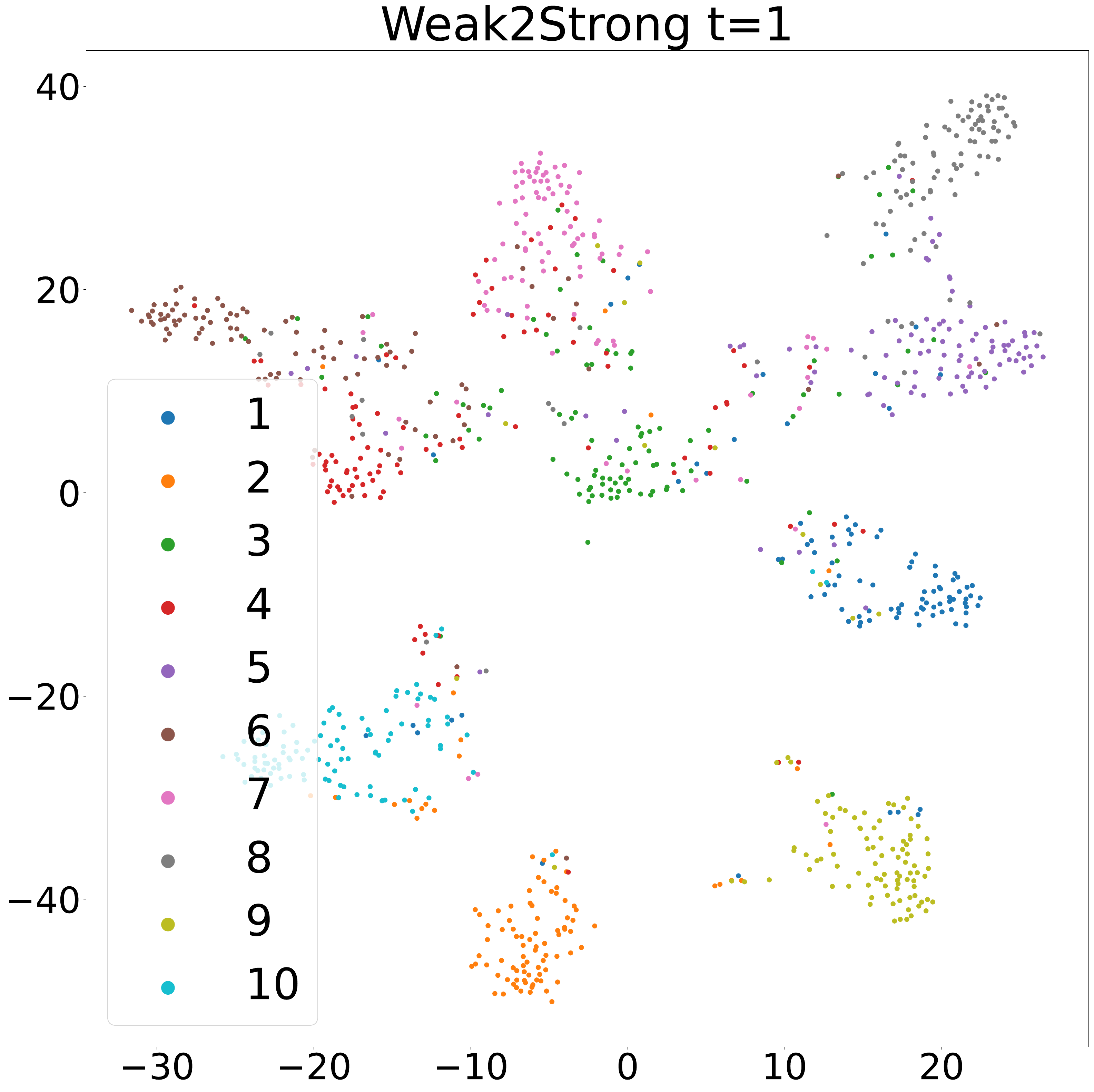}\hspace{-1mm}
	\includegraphics[width=0.2\textwidth]{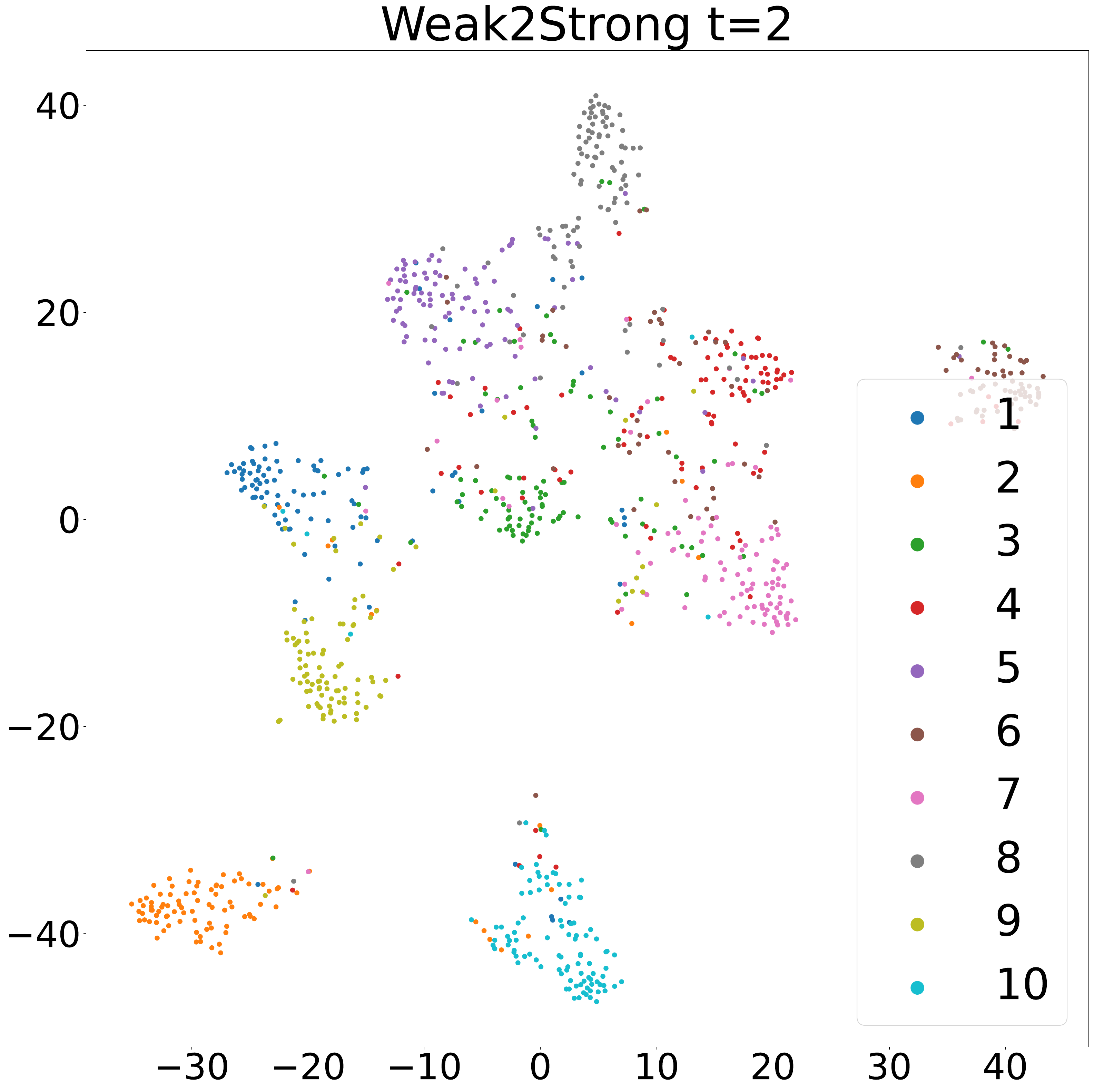}\hspace{-1mm}
	\includegraphics[width=0.2\textwidth]{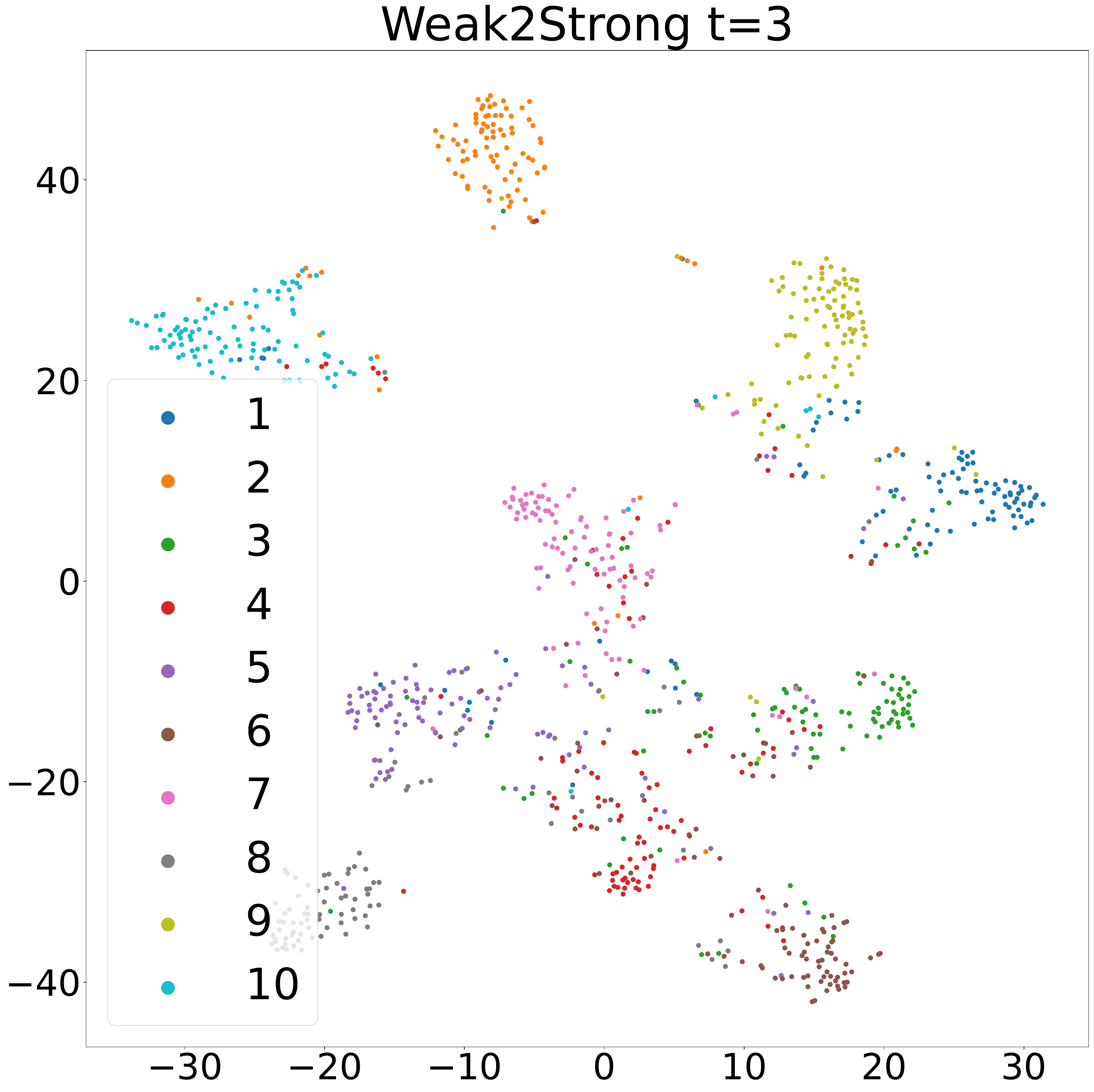}\hspace{-1mm}
	\includegraphics[width=0.2\textwidth]{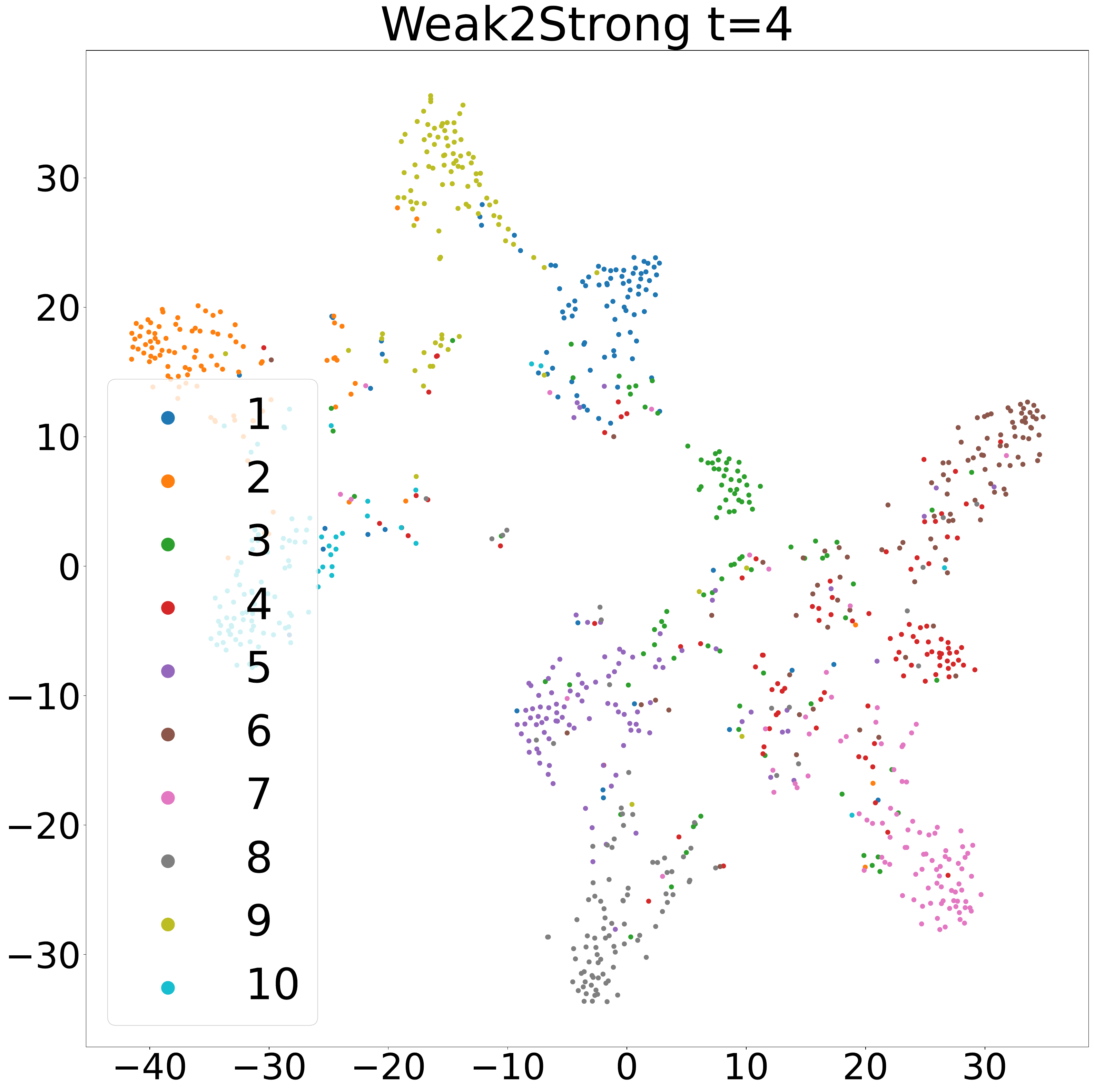}
	}
\vskip -0.05in
\caption{Visualization of the output distribution of each timestep submodel. (a) The vanilla SNN produces confusing outputs at $t=0$, showing dramatic gaps between the strong and the weak. (b) (c) Both Strong2Weak and Weak2Strong are able to improve the output discriminability of the submodels, thus bridging the gap between strong and weak to improve the overall stability and performance.}
\label{fig:tsne}
\vskip -0.15in
\end{figure}

\subsection{Submodel Output Distribution Visualization}
To illustrate more intuitively how the strong and weak submodels enhance each other, the output 2D t-distributed stochastic neighbor embedding (t-SNE) distributions of the submodels for each timestep are visualized in Fig.~\ref{fig:tsne}. In Fig.~\ref{fig:tsne}(a), the vanilla SNN exhibits a large output gap across timesteps and is particularly weak at $t=0$. Excessive differences between the strong and the weak affect the overall performance. Compared to the vanilla SNN, the Strong2Weak distillation guides the weak and compensates the deficiencies, thus contributing to the overall performance. Specifically, Weak2Strong distills the weak into the strong and can improve the performance of the weak as well, embodying the concept of teaching and learning. From an overall perspective, Strong2Weak and Weak2Strong improve the discriminability of the submodel outputs at each timestep, which improves the overall stability and performance. The visualization of the overall output can be found in \textbf{Appendix~\ref{app_C}}.

\subsection{Improved Performance at Low Timesteps}

Timestep and inference latency are positively correlated. If lower latency is required, inference can be performed with fewer timesteps. The performance of the SNN at different inference timesteps is shown in Table~\ref{com_lowtimestep}, using the pre-trained model with 5 timesteps. The performance of the vanilla SNN degrades significantly as the timestep is reduced. In particular, at 1 timestep, the corresponding submodel is too weak, leading to a performance equivalent to a random guess. Membrane potential smoothing~\cite{MPS} mitigates this problem to some extent, and decent performance can be achieved at low latencies. Our distillation schemes narrow the gap in output across timesteps, improving overall stability and maintaining satisfactory performance even with reduced timesteps. This makes it possible to train a model to apply under various latency constraints without retraining.
\begin{table}[t]
  \begin{minipage}{0.49\textwidth}
  \captionof{table}{Comparative results (\%) for reduced timestep inference on CIFAR10-DVS. Our distillation schemes can maintain decent performance even with reduced inference timesteps.}
  \label{com_lowtimestep}
  \tabcolsep=0.025\columnwidth
  \scalebox{0.84}{
  \begin{tabular}{cccccc}
    \toprule
    Method & T=1 & T=2 & T=3 & T=4 & T=5\\
    \midrule
    \rowcolor{gray!8}Vanilla SNN & 10.00 & 60.10 & 69.50 & 73.30 & 74.10\\
    \rowcolor{gray!8}MPS~\cite{MPS} & 66.60 & 74.30 & 75.50 & 75.70 & 76.60\\
    \hline
    \rowcolor{S2Wcolor}Strong2Weak & 71.50 & 76.20 & 77.10 & 78.20 & 78.60\\
    \rowcolor{W2Scolor}Weak2Strong & 73.40 & 76.50 & 77.50 & 78.80 & 79.70\\
    \bottomrule
  \end{tabular}
  }
  \end{minipage}
  \hspace{0.01\textwidth}%
  \begin{minipage}{0.49\textwidth}
  \captionof{table}{Improved adversarial robustness performance (\%) of the proposed distillation schemes. Experiments were performed with VGG-11 on CIFAR100, $T=8$.}
  \label{com_robust}
  \scalebox{0.84}{
  \tabcolsep=0.005\columnwidth
\begin{tabular}{ccccccc}
  \toprule
  Method  & Clean & GN & FGSM & PGD & BIM & CW\\
  \midrule
  FEEL-SNN~\cite{xu2024feelsnn} & 63.95 & 61.97 & 9.87 & 2.13 & 1.93 & 6.21\\
  AT~\cite{goodfellow2014explaining} &  67.97 & 67.47 & 17.55 & 9.52 & 8.91 & 20.23\\
  RAT~\cite{ding2022snnrat} & 69.99 & 69.06 & 19.00 & 9.11 & 8.46 & 22.59\\
  \hline
  \rowcolor{S2Wcolor}\textbf{Strong2Weak} & \textbf{70.68} & \textbf{69.53} & 21.15 & 10.30 & 9.40 & 23.40\\
  \cline{2-7}
  \rowcolor{W2Scolor}\textbf{Weak2Strong} & 70.09 & 68.87 & \textbf{21.23} & \textbf{10.70} & \textbf{9.75} & \textbf{24.08}\\
  \bottomrule
 \end{tabular}
  }
  \end{minipage}
\vspace{-0.2in}
\end{table}

\vspace{-0.1in}
\subsection{Robustness Gains from Self-Distillation}
\label{exp_robust}
\vspace{-0.05in}
The proposed self-distillation schemes improve the overall stability and performance of the SNN by guiding the strong and weak submodels towards each other. Since stability is closely related to robustness~\cite{xu2024feelsnn}, we investigate whether our methods can improve the robustness of SNNs and thus increase the reliability of deployment. We perform robustness experiments on CIFAR100 using VGG-11, and the attack methods include common noise attack (Gaussian noise, GN) and adversarial attacks (FGSM~\cite{goodfellow2014explaining}, PGD with random start~\cite{madry2017towards}, BIM~\cite{kurakin2018adversarial}, and CW~\cite{7958570}). We combine the proposed methods with RAT~\cite{ding2022snnrat} to verify that they can maximize robustness in conjunction with each other. Experimental details are given in the \textbf{Appendix~\ref{app_B}}. 

The comparative adversarial robustness results are shown in Table~\ref{com_robust}, where our method further improves the performance of RAT~\cite{ding2022snnrat} for both clean and adversarial samples. In particular, Weak2Strong improves robust accuracy under FGSM attacks by 2.23\%, which is quite significant. This confirms that the stability provided by self-distillation can indeed improve the robustness of the model against attacks, and we will explore further effects of distillation on robustness promotion in the future.

\vspace{-0.1in}
\subsection{Comparison with Other Methods}
\vspace{-0.05in}

\begin{table}[t]
  \begin{minipage}{0.45\textwidth}
  \caption{Comparative results on ImageNet.}
  \label{com_imagenet}
  \tabcolsep=0.01\columnwidth
  \scalebox{0.84}{
  \begin{tabular}{p{0.01in}cccc}
  \toprule
  & Method  & Architecture & T & Acc.(\%)\\
  \midrule
  \multirow{6}{*}{\rotatebox{90}{\textbf{w/o KD}}}
  & TAB~\cite{TAB}$^{ICLR'24}$ & ResNet34 & 4 & 67.78\\
  & Shortcut~\cite{guo2024take}$^{NeurIPS'24}$ & ResNet34 & 4 & 68.14\\
  & FSTA-SNN~\cite{FSTA-SNN}$^{AAAI'25}$ & ResNet34 & 4 & 70.23\\
  & STAA-SNN~\cite{zhang2025staa}$^{CVPR'25}$ & ResNet34 & 4 & 70.40\\
  & SEW-ResNet~\cite{SEWResNet}$^{NeurIPS'21}$ & SEW-ResNet34 & 4 & 67.04\\
  & IMP+LTS~\cite{IMP}$^{ICLR'25}$ & SEW-ResNet34 & 4 & 68.90\\
  \hline
  \multirow{5}{*}{\rotatebox{90}{\textbf{KD}}}
  & SSCL~\cite{zhang2024enhancing}$^{AAAI'24}$ & ResNet34 & 4 & 66.78\\
  & EnOF~\cite{guo2024enofsnn}$^{NeurIPS'24}$ &  ResNet34 & 4 & 67.40\\
  & KDSNN~\cite{KDSNN}$^{CVPR'23}$ &  SEW-ResNet34 & 4 & 67.28\\
  & MPS~\cite{MPS}$^{ICLR'25}$ & SEW-ResNet34 & 4 & 69.03\\
   & TKS~\cite{TKS}$^{IEEE~TAI'24}$ &  SEW-ResNet34 & 4 & 69.60\\
  \hline
  \rowcolor{S2Wcolor} & \textbf{Strong2Weak} & SEW-ResNet34 & 4 & \textbf{70.53}\\
  \cline{1-5}
  \rowcolor{W2Scolor}  & \textbf{Weak2Strong} & SEW-ResNet34 & 4 & 69.87\\
  \bottomrule
 \end{tabular}
  }
  \end{minipage}
  \hspace{0.07\textwidth}%
  \begin{minipage}{0.48\textwidth}
  \caption{Comparative results on CIFAR10/100.}
  \label{com_cifar}
  \scalebox{0.75}{
  \tabcolsep=0.01\columnwidth
\begin{tabular}{p{0.07in}ccccc}
  \toprule
  & Method  & Architecture & T & CIFAR10 & CIFAR100\\
  \midrule
  \multirow{6}{*}{\rotatebox{90}{\textbf{w/o KD}}}
 & RMP-Loss~\cite{RMPloss} &  VGG-16 & 10 & 94.39 & 73.30\\
 & CLIF~\cite{CLIF} &  ResNet-18 & 4 & 94.89 & 77.00\\
 & NDOT$_O$~\cite{NDOT} & VGG-11 & 4 & 94.79 & 76.18\\
 & MS-ResNet~\cite{MSResNet} &  MS-ResNet18 & 4 & 94.88 & 76.33\\
 & Spikformer~\cite{Spikformer} & Transformer & 4 & 93.94 & 75.96\\
 &  QKFormer~\cite{QKFormer} & QKFormer & 4 & 96.18 & 81.15\\
  \hline
  \multirow{5}{*}{\rotatebox{90}{\textbf{KD}}}
 & SSCL~\cite{zhang2024enhancing} & ResNet-20 & 4 & 94.27 & 72.86\\
 & EnOF~\cite{guo2024enofsnn} & ResNet-20 & 4 & 94.74 & 73.01\\
 & KDSNN~\cite{KDSNN} & ResNet-18 & 4 & 93.41 & - \\
 & BKDSNN~\cite{BKDSNN} & ResNet-19 & 4 & 94.64 & 74.95\\
 & TKS~\cite{TKS} & ResNet-19 & 4 & 96.35 & 79.89\\
  \hline
\rowcolor{S2Wcolor} &  & MS-ResNet18 & 4 & 95.15 & 78.25\\ 
\rowcolor{S2Wcolor}& & ResNet-19 & 4 & \textbf{96.62} & \textbf{81.83}\\
\rowcolor{S2Wcolor}& \multirow{-3}{*}{\textbf{Strong2Weak}}& QKFormer & 4 & 96.42 & 81.27\\
  \cline{1-6}
\rowcolor{W2Scolor} & & MS-ResNet18 & 4 & 95.13 & 77.98\\
\rowcolor{W2Scolor} & & ResNet-19 & 4 & \textbf{96.66} & \textbf{82.02}\\
\rowcolor{W2Scolor} & \multirow{-3}{*}{\textbf{Weak2Strong}} & QKFormer & 4 & 96.29 & 81.23\\
  \bottomrule
 \end{tabular}
  }
  \end{minipage}
\vspace{-0.2in}
\end{table}

\begin{wraptable}[22]{r}{7.5cm}
\vspace{-0.25in}
  \caption{Comparative results (\%) on neuromorphic datasets.}
  \label{com_dvs}
  \vspace{-0.05in}
  \tabcolsep=0.005\columnwidth
  \scalebox{0.8}{
\begin{tabular}{p{0.05in}ccccc}
  \toprule
 &  Method  & Archi. & T & CIFAR10-DVS & DVS-Gesture\\
  \midrule
  \multirow{6}{*}{\rotatebox{90}{\textbf{w/o KD}}}
  & DSR~\cite{DSR} & VGG-11 & 20 & 77.27 & -\\
  & RMP-Loss~\cite{RMPloss} &  ResNet-20 & 10 & 75.60 & -\\
  & NDOT~\cite{NDOT} & VGG-11 & 10 & 77.50 & -\\
  & TAB~\cite{TAB} &  VGG-9 & 5 & 74.57 & 90.86\\
  & SLT~\cite{anumasa2024enhancing} &  VGG-9 & 5 & 74.23 & 89.35\\
  & SSNN~\cite{SSNN} & VGG-9 & 5 & 73.63 & 90.74\\
  \hline
  \multirow{5}{*}{\rotatebox{90}{\textbf{KD}}}
  & MPS~\cite{MPS} & VGG-9 & 5 & 76.77 & 93.23\\
  & TSSD~\cite{TSSD} & VGG-9 & 5 & 72.90 & 86.69\\
  & SSCL~\cite{zhang2024enhancing} & ResNet-20 & 10 & 78.50 & -\\
  & EnOF~\cite{guo2024enofsnn} &  ResNe-20 & 10 & 80.50 & -\\
  & TKS~\cite{TKS} & VGGSNN & 10 & 85.30 & -\\
  \hline
  \rowcolor{S2Wcolor} & & VGG-9 & 5 & 78.93 & 91.43\\
  \rowcolor{S2Wcolor} & \multirow{-2}{*}{\textbf{Strong2Weak}} &VGGSNN & 10 & 85.60 & - \\
  \cline{1-6}
    \rowcolor{W2Scolor} & & VGG-9 & 5 & 79.33 & 91.20\\
  \rowcolor{W2Scolor} & \multirow{-2}{*}{\textbf{Weak2Strong}} &VGGSNN & 10 & \textbf{86.70} & - \\
  \hline
  & SDT~\cite{SDT} & Transformer & 5 & 72.53\tnote{*} & 93.98\tnote{*} \\
  & QKFormer~\cite{QKFormer} & QKFormer & 16 & 84.00 & 98.60 \\
  \hline
  \rowcolor{S2Wcolor} &  &Transformer & 5 & 73.57 & 95.14\\
  \rowcolor{S2Wcolor} & & & 5 & 82.50 & 95.14\\
  \rowcolor{S2Wcolor}  & \multirow{-3}{*}{\textbf{Strong2Weak}} & \multirow{-2}{*}{QKFormer} & 16 & 84.60 & \textbf{99.08}\\
  \cline{1-6}
  \rowcolor{W2Scolor} &  &Transformer & 5 & 73.20 & 94.91\\
  \rowcolor{W2Scolor} & & & 5 & 82.80 & 94.91\\
  \rowcolor{W2Scolor}  & \multirow{-3}{*}{\textbf{Weak2Strong}} & \multirow{-2}{*}{QKFormer} & 16 & 84.90 & 98.96\\
  \bottomrule
 \end{tabular}
 }
\end{wraptable}

\textbf{Static datasets.} The comparative results on ImageNet are shown in Table~\ref{com_imagenet}. Our method achieves 70.53\% accuracy at 4 timesteps, surpassing other distillation methods and specially designed modules. Compared to TKS~\cite{TKS}, which does not require a teacher model for distillation, we outperform it by 0.93\% in accuracy. Additionally, we also achieved accuracies of 96.66\% and 82.02\% on CIFAR10 and CIFAR100, respectively, which again exceeded the other methods, as shown in Table~\ref{com_cifar}. We also show results on Tiny-ImageNet in Table~\ref{com_tinyimagenet}, where our method can be combined with CILF~\cite{CLIF} neurons to further improve performance.

\textbf{Neuromorphic datasets.} The comparative results on neuromorphic datasets are shown in Table~\ref{com_dvs}. Weak2Strong achieved 86.70\% accuracy on CIFAR10-DVS using the VGGSNN architecture with $T=10$, which was 1.40\% higher than TKS~\cite{TKS}. In addition, our distillation schemes also enhance the SNN with Transformer architecture. The performance of both the SDT~\cite{SDT} and the QKFormer~\cite{QKFormer} architectures improves further, demonstrating the versatility of our method.

\subsection{Combined with the Early Exit Dynamic Inference}

Table~\ref{com_early_exit} shows the performance of our method when combined with the early exit mechanism~\cite{SEENN} for dynamic inference. We used the VGG-9 model on the CIFAR10-DVS and DVS-Gesture datasets, setting the early exit thresholds to 0.95, 0.9, 0.8, and 0.5. The results suggest that our method is highly compatible with the early-exit mechanism. Specifically, when the exit threshold is set to 0.8, our method achieves accuracies of 77.8\% (Strong2Weak, $\overline{T}=2.3$) and 78.8\% (Weak2Strong, $\overline{T}=2.27$) on CIFAR10-DVS. Compared to the vanilla SNN, our method demonstrates significantly improved performance while reducing latency by half. This further demonstrates the potential of our method for ultra-low-latency inference.

\begin{table}[t]
  \begin{minipage}{0.49\textwidth}
  \caption{The proposed method can be combined with the early exit dynamic inference. $\vartheta_{exit}$ is the early exit threshold, $\overline{T}$ is the average inference timestep.}
  \label{com_early_exit}
  \tabcolsep=0.01\columnwidth
  \scalebox{0.8}{
  \begin{tabular}{ccc|cc}
  \toprule
 \multirow{2}{*}{Method} & \multicolumn{2}{c}{CIFAR10-DVS} & \multicolumn{2}{c}{DVS-Gesture}\\ & $\overline{T}$ & Acc. (\%) & $\overline{T}$ & Acc. (\%)\\
  \midrule
  Vanilla SNN & 5 & 73.97 & 5 & 87.85\\
  \hline
  \rowcolor{S2Wcolor}Strong2Weak & 5 & 78.60 & 5 & 92.01\\
  \rowcolor{S2Wcolor}+Early exit ($\vartheta_{exit}=0.95$) & 3.16 & 78.40 & 2.35 & 91.32\\
  \rowcolor{S2Wcolor}+Early exit ($\vartheta_{exit}=0.90$) & 2.73 & 78.20 & 1.87 & 90.62\\
  \rowcolor{S2Wcolor}+Early exit ($\vartheta_{exit}=0.80$) & 2.30 & 77.80 & 1.56 & 90.28\\
  \rowcolor{S2Wcolor}+Early exit ($\vartheta_{exit}=0.50$) & 1.38 & 74.70 & 1.06 & 88.54\\
  \hline
  \rowcolor{W2Scolor}Weak2Strong & 5 & 79.70 & 5 & 91.67\\
  \rowcolor{W2Scolor}+Early exit ($\vartheta_{exit}=0.95$) & 3.19 & 79.50 & 2.22 & 91.67\\
  \rowcolor{W2Scolor}+Early exit ($\vartheta_{exit}=0.90$) & 2.74 & 79.40 & 1.89 & 90.62\\
  \rowcolor{W2Scolor}+Early exit ($\vartheta_{exit}=0.80$) & 2.27 & 78.80 & 1.58 & 89.93\\
  \rowcolor{W2Scolor}+Early exit ($\vartheta_{exit}=0.50$) & 1.38 & 75.80 & 1.01 & 86.11\\
  \bottomrule
 \end{tabular}}
  \end{minipage}
  \hspace{0.01\textwidth}%
  \begin{minipage}{0.49\textwidth}
  \caption{The proposed method can be combined with the early exit dynamic inference. $\vartheta_{exit}$ is the early exit threshold, $\overline{T}$ is the average inference timestep.}
  \label{com_metric}
  \scalebox{0.88}{
  \tabcolsep=0.005\columnwidth
\begin{tabular}{cccc}
  \toprule
 Metric & Distillation & CIFAR10-DVS & DVS-Gesture\\
  \midrule
  \rowcolor{S2Wcolor}Confidence & Strong2Weak & 78.93 & 91.43\\
  \rowcolor{S2Wcolor}Entropy & Strong2Weak & 78.80 & 90.97\\
  \rowcolor{S2Wcolor}Margin & Strong2Weak & 78.83 & 91.55\\
  \rowcolor{S2Wcolor}Diversity & Strong2Weak & 78.27 & 90.57\\
  \hline
  \rowcolor{W2Scolor}Confidence & Strong2Weak & 79.33 & 91.20\\
  \rowcolor{W2Scolor}Entropy & Strong2Weak & 78.57 & 90.05\\
  \rowcolor{W2Scolor}Margin & Strong2Weak & 78.57 & 90.97\\
  \rowcolor{W2Scolor}Diversity & Strong2Weak & 78.73 & 91.78\\
  \bottomrule
 \end{tabular}}
  \end{minipage}
\vspace{-0.2in}
\end{table}

\subsection{The Influence of Metrics on the Strong and the Weak}

We use confidence as the default metric to distinguish between strong and weak submodels because it can be efficiently calculated using only a softmax function. In addition, confidence typically reflects the uncertainty of model output results and has been shown to be related to generalization error, which to a certain extent reflects performance~\cite{6203595}. Due to these advantages, confidence is widely used to evaluate model performance in areas such as model calibration~\cite{pmlr-v70-guo17a,pmlr-v162-wei22d} and early exit~\cite{Yang_2020_CVPR,SEENN}.

As a general distillation method, we can seamlessly replace confidence with other metrics such as entropy, margin (maximum confidence difference between target category and other categories), diversity (confidence differences between all categories), or other specially designed metrics. Table~\ref{com_metric} shows the influence of different metrics on performance. The results suggest that different metrics do not yield significant performance differences. This indicates that the straightforward and intuitive confidence is sufficient and effective for distinguishing between strong and weak submodels.

\vspace{-0.1in}
\section{Conclusion}

In this paper, we deconstruct the multi timestep SNN into submodels and distill between the submodels, thereby eliminating the need for additional overhead. We identify the strong and the weak by evaluating the output confidence of the submodels, and propose two self-distillation schemes, Strong2Weak and Weak2Strong, respectively, where the strong helps the weak and the weak transfers the underlying dark knowledge to the strong. These two distillation schemes can be flexibly implemented, such as one-to-one, ensemble, simultaneous, and cascade distillation, with great extensibility. Extensive visualizations and experiments show that this efficient self-distillation method effectively mitigates strong and weak gaps between submodels and significantly improves low-latency inference performance. In particular, the adversarial robustness of the model is also gained with the overall stability improvement delivered by self-distillation. This will contribute to efficient and high-performance studies of SNNs.

\textbf{Limitation.} The aim of this paper is to demonstrate efficient self-distillation schemes for SNNs, and therefore there is no deliberate tuning of the distillation loss functions and coefficients, which allows the performance of the proposed method to be further enhanced by improving these.

Any highly efficient, high-performance model is at risk of misuse in the real world, especially SNNs, which are known for their low power consumption and low latency. To mitigate the risk of misuse, we believe it is necessary to integrate offensive and defensive strategies and security measures into the proposed method. As an alternative, privacy-preserving mechanisms, such as federated learning, should be incorporated during deployment.

\section*{Acknowledgments}

This work was supported by the National Natural Science Foundation of China under Grant No. 62276054 and 62406060.

\medskip
{
\small
\bibliographystyle{unsrt}
\bibliography{ref}
}


\newpage
\appendix

\newpage
\section*{NeurIPS Paper Checklist}

\begin{enumerate}

\item {\bf Claims}
    \item[] Question: Do the main claims made in the abstract and introduction accurately reflect the paper's contributions and scope?
    \item[] Answer: \answerYes{} 
    \item[] Justification: The abstract and introduction accurately reflect the contribution and scope of this paper.
    \item[] Guidelines:
    \begin{itemize}
        \item The answer NA means that the abstract and introduction do not include the claims made in the paper.
        \item The abstract and/or introduction should clearly state the claims made, including the contributions made in the paper and important assumptions and limitations. A No or NA answer to this question will not be perceived well by the reviewers. 
        \item The claims made should match theoretical and experimental results, and reflect how much the results can be expected to generalize to other settings. 
        \item It is fine to include aspirational goals as motivation as long as it is clear that these goals are not attained by the paper. 
    \end{itemize}

\item {\bf Limitations}
    \item[] Question: Does the paper discuss the limitations of the work performed by the authors?
    \item[] Answer: \answerYes{} 
    \item[] Justification: We explicitly point out the limitations of this paper.
    \item[] Guidelines:
    \begin{itemize}
        \item The answer NA means that the paper has no limitation while the answer No means that the paper has limitations, but those are not discussed in the paper. 
        \item The authors are encouraged to create a separate "Limitations" section in their paper.
        \item The paper should point out any strong assumptions and how robust the results are to violations of these assumptions (e.g., independence assumptions, noiseless settings, model well-specification, asymptotic approximations only holding locally). The authors should reflect on how these assumptions might be violated in practice and what the implications would be.
        \item The authors should reflect on the scope of the claims made, e.g., if the approach was only tested on a few datasets or with a few runs. In general, empirical results often depend on implicit assumptions, which should be articulated.
        \item The authors should reflect on the factors that influence the performance of the approach. For example, a facial recognition algorithm may perform poorly when image resolution is low or images are taken in low lighting. Or a speech-to-text system might not be used reliably to provide closed captions for online lectures because it fails to handle technical jargon.
        \item The authors should discuss the computational efficiency of the proposed algorithms and how they scale with dataset size.
        \item If applicable, the authors should discuss possible limitations of their approach to address problems of privacy and fairness.
        \item While the authors might fear that complete honesty about limitations might be used by reviewers as grounds for rejection, a worse outcome might be that reviewers discover limitations that aren't acknowledged in the paper. The authors should use their best judgment and recognize that individual actions in favor of transparency play an important role in developing norms that preserve the integrity of the community. Reviewers will be specifically instructed to not penalize honesty concerning limitations.
    \end{itemize}

\item {\bf Theory assumptions and proofs}
    \item[] Question: For each theoretical result, does the paper provide the full set of assumptions and a complete (and correct) proof?
    \item[] Answer: \answerNA{} 
    \item[] Justification: We demonstrate the superior performance of our method through experimentation.
    \item[] Guidelines:
    \begin{itemize}
        \item The answer NA means that the paper does not include theoretical results. 
        \item All the theorems, formulas, and proofs in the paper should be numbered and cross-referenced.
        \item All assumptions should be clearly stated or referenced in the statement of any theorems.
        \item The proofs can either appear in the main paper or the supplemental material, but if they appear in the supplemental material, the authors are encouraged to provide a short proof sketch to provide intuition. 
        \item Inversely, any informal proof provided in the core of the paper should be complemented by formal proofs provided in appendix or supplemental material.
        \item Theorems and Lemmas that the proof relies upon should be properly referenced. 
    \end{itemize}

    \item {\bf Experimental result reproducibility}
    \item[] Question: Does the paper fully disclose all the information needed to reproduce the main experimental results of the paper to the extent that it affects the main claims and/or conclusions of the paper (regardless of whether the code and data are provided or not)?
    \item[] Answer: \answerYes{} 
    \item[] Justification: We clearly illustrate the experimental details.
    \item[] Guidelines:
    \begin{itemize}
        \item The answer NA means that the paper does not include experiments.
        \item If the paper includes experiments, a No answer to this question will not be perceived well by the reviewers: Making the paper reproducible is important, regardless of whether the code and data are provided or not.
        \item If the contribution is a dataset and/or model, the authors should describe the steps taken to make their results reproducible or verifiable. 
        \item Depending on the contribution, reproducibility can be accomplished in various ways. For example, if the contribution is a novel architecture, describing the architecture fully might suffice, or if the contribution is a specific model and empirical evaluation, it may be necessary to either make it possible for others to replicate the model with the same dataset, or provide access to the model. In general. releasing code and data is often one good way to accomplish this, but reproducibility can also be provided via detailed instructions for how to replicate the results, access to a hosted model (e.g., in the case of a large language model), releasing of a model checkpoint, or other means that are appropriate to the research performed.
        \item While NeurIPS does not require releasing code, the conference does require all submissions to provide some reasonable avenue for reproducibility, which may depend on the nature of the contribution. For example
        \begin{enumerate}
            \item If the contribution is primarily a new algorithm, the paper should make it clear how to reproduce that algorithm.
            \item If the contribution is primarily a new model architecture, the paper should describe the architecture clearly and fully.
            \item If the contribution is a new model (e.g., a large language model), then there should either be a way to access this model for reproducing the results or a way to reproduce the model (e.g., with an open-source dataset or instructions for how to construct the dataset).
            \item We recognize that reproducibility may be tricky in some cases, in which case authors are welcome to describe the particular way they provide for reproducibility. In the case of closed-source models, it may be that access to the model is limited in some way (e.g., to registered users), but it should be possible for other researchers to have some path to reproducing or verifying the results.
        \end{enumerate}
    \end{itemize}

\item {\bf Open access to data and code}
    \item[] Question: Does the paper provide open access to the data and code, with sufficient instructions to faithfully reproduce the main experimental results, as described in supplemental material?
    \item[] Answer: \answerYes{} 
    \item[] Justification: Our method is based on open-source datasets and existing methods, and provides detailed experimental details.
    \item[] Guidelines:
    \begin{itemize}
        \item The answer NA means that paper does not include experiments requiring code.
        \item Please see the NeurIPS code and data submission guidelines (\url{https://nips.cc/public/guides/CodeSubmissionPolicy}) for more details.
        \item While we encourage the release of code and data, we understand that this might not be possible, so “No” is an acceptable answer. Papers cannot be rejected simply for not including code, unless this is central to the contribution (e.g., for a new open-source benchmark).
        \item The instructions should contain the exact command and environment needed to run to reproduce the results. See the NeurIPS code and data submission guidelines (\url{https://nips.cc/public/guides/CodeSubmissionPolicy}) for more details.
        \item The authors should provide instructions on data access and preparation, including how to access the raw data, preprocessed data, intermediate data, and generated data, etc.
        \item The authors should provide scripts to reproduce all experimental results for the new proposed method and baselines. If only a subset of experiments are reproducible, they should state which ones are omitted from the script and why.
        \item At submission time, to preserve anonymity, the authors should release anonymized versions (if applicable).
        \item Providing as much information as possible in supplemental material (appended to the paper) is recommended, but including URLs to data and code is permitted.
    \end{itemize}

\item {\bf Experimental setting/details}
    \item[] Question: Does the paper specify all the training and test details (e.g., data splits, hyperparameters, how they were chosen, type of optimizer, etc.) necessary to understand the results?
    \item[] Answer: \answerYes{} 
    \item[] Justification: We provide the details of the experiments in the Appendix.
    \item[] Guidelines:
    \begin{itemize}
        \item The answer NA means that the paper does not include experiments.
        \item The experimental setting should be presented in the core of the paper to a level of detail that is necessary to appreciate the results and make sense of them.
        \item The full details can be provided either with the code, in appendix, or as supplemental material.
    \end{itemize}

\item {\bf Experiment statistical significance}
    \item[] Question: Does the paper report error bars suitably and correctly defined or other appropriate information about the statistical significance of the experiments?
    \item[] Answer: \answerYes{} 
    \item[] Justification: We report the average results of three trials to minimize the errors.
    \item[] Guidelines:
    \begin{itemize}
        \item The answer NA means that the paper does not include experiments.
        \item The authors should answer "Yes" if the results are accompanied by error bars, confidence intervals, or statistical significance tests, at least for the experiments that support the main claims of the paper.
        \item The factors of variability that the error bars are capturing should be clearly stated (for example, train/test split, initialization, random drawing of some parameter, or overall run with given experimental conditions).
        \item The method for calculating the error bars should be explained (closed form formula, call to a library function, bootstrap, etc.)
        \item The assumptions made should be given (e.g., Normally distributed errors).
        \item It should be clear whether the error bar is the standard deviation or the standard error of the mean.
        \item It is OK to report 1-sigma error bars, but one should state it. The authors should preferably report a 2-sigma error bar than state that they have a 96\% CI, if the hypothesis of Normality of errors is not verified.
        \item For asymmetric distributions, the authors should be careful not to show in tables or figures symmetric error bars that would yield results that are out of range (e.g. negative error rates).
        \item If error bars are reported in tables or plots, The authors should explain in the text how they were calculated and reference the corresponding figures or tables in the text.
    \end{itemize}

\item {\bf Experiments compute resources}
    \item[] Question: For each experiment, does the paper provide sufficient information on the computer resources (type of compute workers, memory, time of execution) needed to reproduce the experiments?
    \item[] Answer: \answerYes{} 
    \item[] Justification: We provide information about the computing platform.
    \item[] Guidelines:
    \begin{itemize}
        \item The answer NA means that the paper does not include experiments.
        \item The paper should indicate the type of compute workers CPU or GPU, internal cluster, or cloud provider, including relevant memory and storage.
        \item The paper should provide the amount of compute required for each of the individual experimental runs as well as estimate the total compute. 
        \item The paper should disclose whether the full research project required more compute than the experiments reported in the paper (e.g., preliminary or failed experiments that didn't make it into the paper). 
    \end{itemize}
    
\item {\bf Code of ethics}
    \item[] Question: Does the research conducted in the paper conform, in every respect, with the NeurIPS Code of Ethics \url{https://neurips.cc/public/EthicsGuidelines}?
    \item[] Answer: \answerYes{} 
    \item[] Justification:  Our research adheres to the NeurIPS Code of Ethics in every respect.
    \item[] Guidelines:
    \begin{itemize}
        \item The answer NA means that the authors have not reviewed the NeurIPS Code of Ethics.
        \item If the authors answer No, they should explain the special circumstances that require a deviation from the Code of Ethics.
        \item The authors should make sure to preserve anonymity (e.g., if there is a special consideration due to laws or regulations in their jurisdiction).
    \end{itemize}

\item {\bf Broader impacts}
    \item[] Question: Does the paper discuss both potential positive societal impacts and negative societal impacts of the work performed?
    \item[] Answer: \answerNA{} 
    \item[] Justification: This paper has no potential social impact.
    \item[] Guidelines:
    \begin{itemize}
        \item The answer NA means that there is no societal impact of the work performed.
        \item If the authors answer NA or No, they should explain why their work has no societal impact or why the paper does not address societal impact.
        \item Examples of negative societal impacts include potential malicious or unintended uses (e.g., disinformation, generating fake profiles, surveillance), fairness considerations (e.g., deployment of technologies that could make decisions that unfairly impact specific groups), privacy considerations, and security considerations.
        \item The conference expects that many papers will be foundational research and not tied to particular applications, let alone deployments. However, if there is a direct path to any negative applications, the authors should point it out. For example, it is legitimate to point out that an improvement in the quality of generative models could be used to generate deepfakes for disinformation. On the other hand, it is not needed to point out that a generic algorithm for optimizing neural networks could enable people to train models that generate Deepfakes faster.
        \item The authors should consider possible harms that could arise when the technology is being used as intended and functioning correctly, harms that could arise when the technology is being used as intended but gives incorrect results, and harms following from (intentional or unintentional) misuse of the technology.
        \item If there are negative societal impacts, the authors could also discuss possible mitigation strategies (e.g., gated release of models, providing defenses in addition to attacks, mechanisms for monitoring misuse, mechanisms to monitor how a system learns from feedback over time, improving the efficiency and accessibility of ML).
    \end{itemize}
    
\item {\bf Safeguards}
    \item[] Question: Does the paper describe safeguards that have been put in place for responsible release of data or models that have a high risk for misuse (e.g., pretrained language models, image generators, or scraped datasets)?
    \item[] Answer: \answerNA{} 
    \item[] Justification: This point is irrelevant to the topic of this paper.
    \item[] Guidelines:
    \begin{itemize}
        \item The answer NA means that the paper poses no such risks.
        \item Released models that have a high risk for misuse or dual-use should be released with necessary safeguards to allow for controlled use of the model, for example by requiring that users adhere to usage guidelines or restrictions to access the model or implementing safety filters. 
        \item Datasets that have been scraped from the Internet could pose safety risks. The authors should describe how they avoided releasing unsafe images.
        \item We recognize that providing effective safeguards is challenging, and many papers do not require this, but we encourage authors to take this into account and make a best faith effort.
    \end{itemize}

\item {\bf Licenses for existing assets}
    \item[] Question: Are the creators or original owners of assets (e.g., code, data, models), used in the paper, properly credited and are the license and terms of use explicitly mentioned and properly respected?
    \item[] Answer: \answerYes{} 
    \item[] Justification: We use open-source assets and provide explicit references.
    \item[] Guidelines:
    \begin{itemize}
        \item The answer NA means that the paper does not use existing assets.
        \item The authors should cite the original paper that produced the code package or dataset.
        \item The authors should state which version of the asset is used and, if possible, include a URL.
        \item The name of the license (e.g., CC-BY 4.0) should be included for each asset.
        \item For scraped data from a particular source (e.g., website), the copyright and terms of service of that source should be provided.
        \item If assets are released, the license, copyright information, and terms of use in the package should be provided. For popular datasets, \url{paperswithcode.com/datasets} has curated licenses for some datasets. Their licensing guide can help determine the license of a dataset.
        \item For existing datasets that are re-packaged, both the original license and the license of the derived asset (if it has changed) should be provided.
        \item If this information is not available online, the authors are encouraged to reach out to the asset's creators.
    \end{itemize}

\item {\bf New assets}
    \item[] Question: Are new assets introduced in the paper well documented and is the documentation provided alongside the assets?
    \item[] Answer: \answerNA{} 
    \item[] Justification: This point is irrelevant to the topic of this paper.
    \item[] Guidelines:
    \begin{itemize}
        \item The answer NA means that the paper does not release new assets.
        \item Researchers should communicate the details of the dataset/code/model as part of their submissions via structured templates. This includes details about training, license, limitations, etc. 
        \item The paper should discuss whether and how consent was obtained from people whose asset is used.
        \item At submission time, remember to anonymize your assets (if applicable). You can either create an anonymized URL or include an anonymized zip file.
    \end{itemize}

\item {\bf Crowdsourcing and research with human subjects}
    \item[] Question: For crowdsourcing experiments and research with human subjects, does the paper include the full text of instructions given to participants and screenshots, if applicable, as well as details about compensation (if any)? 
    \item[] Answer: \answerNA{} 
    \item[] Justification: This point is irrelevant to the topic of this paper.
    \item[] Guidelines:
    \begin{itemize}
        \item The answer NA means that the paper does not involve crowdsourcing nor research with human subjects.
        \item Including this information in the supplemental material is fine, but if the main contribution of the paper involves human subjects, then as much detail as possible should be included in the main paper. 
        \item According to the NeurIPS Code of Ethics, workers involved in data collection, curation, or other labor should be paid at least the minimum wage in the country of the data collector. 
    \end{itemize}

\item {\bf Institutional review board (IRB) approvals or equivalent for research with human subjects}
    \item[] Question: Does the paper describe potential risks incurred by study participants, whether such risks were disclosed to the subjects, and whether Institutional Review Board (IRB) approvals (or an equivalent approval/review based on the requirements of your country or institution) were obtained?
    \item[] Answer: \answerNA{} 
    \item[] Justification: This point is irrelevant to the topic of this paper.
    \item[] Guidelines:
    \begin{itemize}
        \item The answer NA means that the paper does not involve crowdsourcing nor research with human subjects.
        \item Depending on the country in which research is conducted, IRB approval (or equivalent) may be required for any human subjects research. If you obtained IRB approval, you should clearly state this in the paper. 
        \item We recognize that the procedures for this may vary significantly between institutions and locations, and we expect authors to adhere to the NeurIPS Code of Ethics and the guidelines for their institution. 
        \item For initial submissions, do not include any information that would break anonymity (if applicable), such as the institution conducting the review.
    \end{itemize}

\item {\bf Declaration of LLM usage}
    \item[] Question: Does the paper describe the usage of LLMs if it is an important, original, or non-standard component of the core methods in this research? Note that if the LLM is used only for writing, editing, or formatting purposes and does not impact the core methodology, scientific rigorousness, or originality of the research, declaration is not required.
    \item[] Answer: \answerNA{} 
    \item[] Justification: This point is irrelevant to the topic of this paper.
    \item[] Guidelines:
    \begin{itemize}
        \item The answer NA means that the core method development in this research does not involve LLMs as any important, original, or non-standard components.
        \item Please refer to our LLM policy (\url{https://neurips.cc/Conferences/2025/LLM}) for what should or should not be described.
    \end{itemize}

\end{enumerate}

\newpage

\section{Appendix A: Flexible Self-Distillation Implementation}
\label{app_A}

While Strong2Weak and Weak2Strong are implemented as one-to-one distillations by default, both can be implemented in flexible ways such as ensemble teacher, ensemble student, simultaneous distillation, and cascade distillation. Below are descriptions of alternative distillation implementations.

\textbf{Ensemble teacher:} For an SNN with $T>2$ timesteps, it can be deconstructed into $T$ submodels. In this case, one submodel is taken as the student, while the remaining set of $T-1$ submodels is ensembled as the teacher model to guide the student. When Strong2Weak is implemented, the student model is the one with the lowest confidence; when Weak2Strong is implemented, the student model is the one with the highest confidence.

\textbf{Ensemble student:} For $T>2$ submodels, the ensemble student takes one of the submodels as the teacher, while the remaining ensemble submodels are distilled as the student model, which is the opposite of the ensemble teacher.

\textbf{Simultaneous distillation:} For the strong submodel with the highest confidence and the weak submodel with the lowest confidence, simultaneous distillation is performed for both Strong2Weak and Weak2Strong distillation, i.e., the final loss is the cross-entropy loss $\mathcal{L}_{CE}(O,Y)$ + $\lambda_{S2W} \mathcal{L}_{S2W}$ + $\lambda_{W2S} \mathcal{L}_{W2S}$. 

\textbf{Cascade distillation:} For $T>2$ submodels, cascade distillation ranks them in order of decreasing confidence to obtain submodels $\{Sub_1,Sub_2,... ,Sub_T\}$, and then distillation is performed between two submodels with adjacent rankings. For example, for $Sub_t$ and $Sub_{t+1}$, the implementation of Strong2Weak takes $Sub_t$ as the teacher and makes $Sub_{t+1}$ the student, and the opposite is true for the implementation of Weak2Strong. We provide PyTorch-style pseudocode for cascade distillation in Alg.~\ref{alg:cascade}.

\begin{algorithm}[t]\small
    \caption{Pseudocode for the proposed self-distillation schemes}
    \label{alg: strong2weak}
    \definecolor{codeblue}{rgb}{0.25,0.5,0.5}
    \definecolor{codepink}{rgb}{1,0.5,0.5}
    \definecolor{codedark}{rgb}{1,0.7,0.8}
    \lstset{
        backgroundcolor=\color{white},
        basicstyle=\fontsize{7.2pt}{7.2pt}\ttfamily\selectfont,
        columns=fullflexible,
        breaklines=true,
        captionpos=b,
        commentstyle=\fontsize{8pt}{8pt}\color{codeblue},
        keywordstyle=\fontsize{8.0pt}{8.0pt}\color{codepink},
        emph={DemoNet, Blk}, %
        emphstyle=\color{purple}, %
    }
    {\small
    \begin{lstlisting}[language=python]
# x: Input data with dimension [T,B,C,H,W].
# y: Label information with dimension [B,M].
# snn: The SNN.
# out: Output of the SNN with dimension [T,B,M].
def self_distillation(x,y,snn,lambda):
    out = snn(x)
    strong,weak = identify(out)
    p_s,p_w = softmax(strong),softmax(weak)
    L_ce = cross_entropy(out.mean(0),y)
    if Strong2Weak:
        L_s2w = KL_div(log(p_w), p_s)
        L_total = L_ce + lambda*L_s2w
    elif Weak2Strong:
        L_w2s = KL_div(log(p_s),p_w)
        L_total = L_ce + lambda*L_w2s
    return L_total
 
# con: Submodel output confidence with dim [T].
def identify(out):
    T = out.shape[0]
    con = zeros(T)
    for t in range(T):
        tmp = max(softmax(out[t], dim=1), dim=1)
        con[t] = mean(tmp)
    max_index,min_index = argmax(con),argmin(con)
    strong = out[max_index]
    weak = out[min_index]
    return strong,weak
    \end{lstlisting}
    }
\end{algorithm}

\begin{algorithm}[t]\small
    \caption{PyTorch-style pseudocode for implementing cascade distillation}
    \label{alg:cascade}
    \definecolor{codeblue}{rgb}{0.25,0.5,0.5}
    \definecolor{codepink}{rgb}{1,0.5,0.5}
    \definecolor{codedark}{rgb}{1,0.7,0.8}
    \lstset{
        backgroundcolor=\color{white},
        basicstyle=\fontsize{7.2pt}{7.2pt}\ttfamily\selectfont,
        columns=fullflexible,
        breaklines=true,
        captionpos=b,
        commentstyle=\fontsize{8pt}{8pt}\color{codeblue},
        keywordstyle=\fontsize{8.0pt}{8.0pt}\color{codepink},
        emph={DemoNet, Blk}, %
        emphstyle=\color{purple}, %
    }
    {\small
    \begin{lstlisting}[language=python]
# x: Input data with dimension [T,B,C,H,W].
# y: Label information with dimension [B,M].
# snn: The SNN.
# out: Output of the SNN with dimension [T,B,M].
def cascade_distillation(x,y,snn,lambda):
    out = snn(x)
    loss = 0
    ranked_indices = rank(out)
    for i in range(T-1):
        strong = out[ranked_indices[i]]
        weak = out[ranked_indices[i+1]]
        p_s,p_w = softmax(strong),softmax(weak)
        if Strong2Weak:
            loss += KL_div(log(p_w),p_s)
        elif Weak2Strong:
            loss += KL_div(log(p_s),p_w)
    loss /= (T-1)
    L_ce = cross_entropy(out.mean(0),y)
    L_total = L_ce + lambda*loss
    return L_total
 
# con: Submodel output confidence with dim [T].
# ranked_indices: Timestep indices ranked in descending confidence order.
def rank(out):
    T = out.shape[0]
    con = zeros(T)
    for t in range(T):
        tmp = max(softmax(out[t], dim=1), dim=1)
        con[t] = mean(tmp)
    ranked_indices = torch.argsort(con, descending=True)
    return ranked_indices
    \end{lstlisting}
    }
\end{algorithm}

\begin{table}[h]
 \centering
  \caption{Performance with flexible distillation implementations (\%). The coefficient for all distillation losses is set to 1. Performance varies slightly for different implementations, with cascade distillation consistently giving better performance.}
 \label{com_implementation}
 \scalebox{0.94}{
 \begin{tabular}{ccc|cc}
  \toprule
 \multirow{2}{*}{Distillation configuration} & \multicolumn{2}{c}{CIFAR10-DVS} & \multicolumn{2}{c}{DVS-Gesture}\\ & S2W & W2S & S2W & W2S\\
  \midrule
  Default & 78.93 & 79.33 & 91.43 & 91.20\\
  Ensemble teacher & 78.53 & 79.17 & 92.02 & 91.78\\
  Ensemble student & 78.30 & 78.37 & 91.55 & 91.55\\
  Simultaneous & \multicolumn{2}{c}{79.17} & \multicolumn{2}{c}{91.32}\\
  Cascade & 79.03 & 79.37 & 93.41 & 92.13\\
  \bottomrule
 \end{tabular}
 }
\end{table}

\subsection{Performance Analysis of Various Implementations} 

The performance of different implementations on CIFAR10-DVS and DVS-Gesture is shown in Table.~\ref{com_implementation}. The results show that the performance of these implementations varies slightly from dataset to dataset, where cascade distillation consistently delivers better performance. In addition, we would like to emphasize that we did not deliberately adjust the hyperparameters in Table.~\ref{com_implementation}, such as the coefficients of Strong2Wweak and Weak2Strong losses, which should have an impact on the performance, especially for simultaneous distillation.

\textbf{Performance Gains from Simultaneous Distillation.} The simultaneous distillation of the weak by the strong and the distillation of the strong by the weak is able to utilize both the representational power of the strong and the dark knowledge of the weak, and thus should be expected to yield greater gains than either alone. However, in Table.~\ref{com_implementation}, simultaneous distillation does not perform optimally as expected. We attribute this to too much similarity between multiple submodels. From an ensemble learning perspective~\cite{9677845,MPS,NEURIPS2020_b86e8d03}, excessive similarity between members leads to a decrease in overall diversity, which is unfavorable for generalization. However, how to balance similarity and diversity is a perennial challenge in ensemble learning~\cite{9677845,NEURIPS2020_b86e8d03} that we leave for future work.

In addition, the adjustment of the two distillation loss coefficients is expected to reduce the conflict between sim ilarity and diversity, thereby improving the performance of simultaneous distillation. For example, incorporating advanced knowledge distillation loss adjustment methods and multi-task loss factor settings. Since our goal is to provide efficient self-distillation schemes for SNNs rather than carefully selecting optimal hyperparameter settings, therefore in this paper we do not deliberately combine other methods to improve the accuracy. When using the self-distillation schemes provided in this paper for performance-critical tasks, incorporating additional strategies to improve performance is further work worth exploring.

\section{Appendix B: Experimental Details}
\label{app_B}

\subsection{Datasets}
We perform experiments on static images and neuromorphic datasets.

CIFAR10 and CIFAR100~\cite{CIFAR} are static image benchmark datasets containing 10 and 100 classes of $32\times32$ color images, respectively. Both datasets contain 50,000 training images and 10,000 test images. For CIFAR10 and CIFAR100 data, we preprocessed them using standard data augmentation strategies: random cropping, horizontal flipping, and normalization. We also use the AutoAugment strategy for CIFAR10.

The ImageNet dataset of 1.2 million training images, 50,000 validation images, and 150,000 test images with 1,000 categories is the most challenging object recognition benchmark. For the ImageNet dataset, we unify the images to a $224\times224$ size during training and testing, and evaluate the performance of our method on the test set.

CIFAR10-DVS dataset~\cite{CIFAR10-DVS} is the neuromorphic version of the CIFAR10 dataset. The CIFAR10-DVS dataset has 10,000 samples for a total of 10 object classes, and the dimension of each sample is $[t,p,x,y]$, where $t$ is the timestamp, $p$ is the polarity of the intensity change of the corresponding pixel, and $x$ and $y$ are the spatial coordinates of the pixel point, respectively. The spatial size of each sample in CIFAR10-DVS is $128\times128$, which we downsampled to $48\times48$ resolution before inputting to the SNN. Additionally, due to the high temporal resolution of the neuromorphic dataset, we integrate a neuromorphic sample into $T$ event frames [$T,p,x,y]$ using the SpikingJelly framework~\cite{SpikingJelly} to match the timestep $T$ of the SNN. For each training, we randomly divide 90\% of the data as the training set and test on the remaining 10\% of the data, which is by far the most common strategy~\cite{SSNN}.

The DVS-Gesture~\cite{DVS-Gesture} dataset contains neuromorphic data for 11 hand gestures with 1176 training samples and 288 test samples. The dimension of each sample is [$T,p,x,y]$, and we downsample its spatial resolution from $128\times128$ to $48\times48$ before feeding the samples into the SNN. The pre-processing of the DVS-Gesture data is the same as in CIFAR10-DVS, which also utilizes the SpikingJelly framework to obtain the event frame [$T,p,x,y]$ by integrating it by timestep.

\subsection{Implementation Details}

Our experiments are based on the PyTorch package running on an Nvidia RTX 4090 GPU. For the VGG-9 and MS-ResNet architectures, we follow the training strategy of~\cite{SSNN}: train the model with an initial learning rate of 0.1 for 100 epochs, reducing it by a factor of ten every 30 epochs. A stochastic gradient descent optimizer with a momentum of 0.9 and a batch size of 64 was used. The weight decays for the static and neuromorphic datasets are 1e-4 and 1e-3, respectively. We used the LIF neuron model with a firing threshold $\vartheta$ of 1.0 and a membrane potential time constant $\tau$ of 2.0.

When using CLIF neurons, we replace the LIF neurons in our model with CLIF neurons and leave the other parameters unchanged. We use the CLIF neuron implementation from the publicly available code of the original paper~\cite{CLIF}.

When using the Spike-driven Transformer architecture, we followed the training strategy of the original model on CIFAR10-DVS~\cite{SDT}: 200 epochs were trained using the Spike-driven Transformer-2-256 architecture, see~\cite{SDT} for details.

When using the ResNet19 architecture, for a fair comparison with TKS~\cite{TKS}, we use the same training strategy as TKS and use their published code.

For the QKFromer~\cite{QKFormer} architecture, we use the proposed method directly on the officially released code and follow its training settings. However, when we used our self-distillation for QKFormer, we noticed a performance degradation when we set the loss function coefficient to 1, so we set this value to 0.01.

To reduce the influence of randomness, we repeated all our experiments three times, and the average results are reported in the paper. Notably, when evaluating the performance of different inference timesteps using the pre-trained five-timestep model, we only report the results of the evaluation on a single pre-trained model, and do not conduct multiple experiments using multiple models.

To evaluate the robustness of the proposed method, we follow the training setup of~\cite{xu2024feelsnn} and rely on its publicly available code. The parameters, architecture, etc. related to the robustness experiment are the same as in~\cite{xu2024feelsnn}. For robustness results, we report the results of a single trial.

\section{Appendix C: Additional Visualizations}
\label{app_C}

\subsection{t-SNE Visualization}
\begin{figure*}[h]
\centering
\includegraphics[width=1.7in]{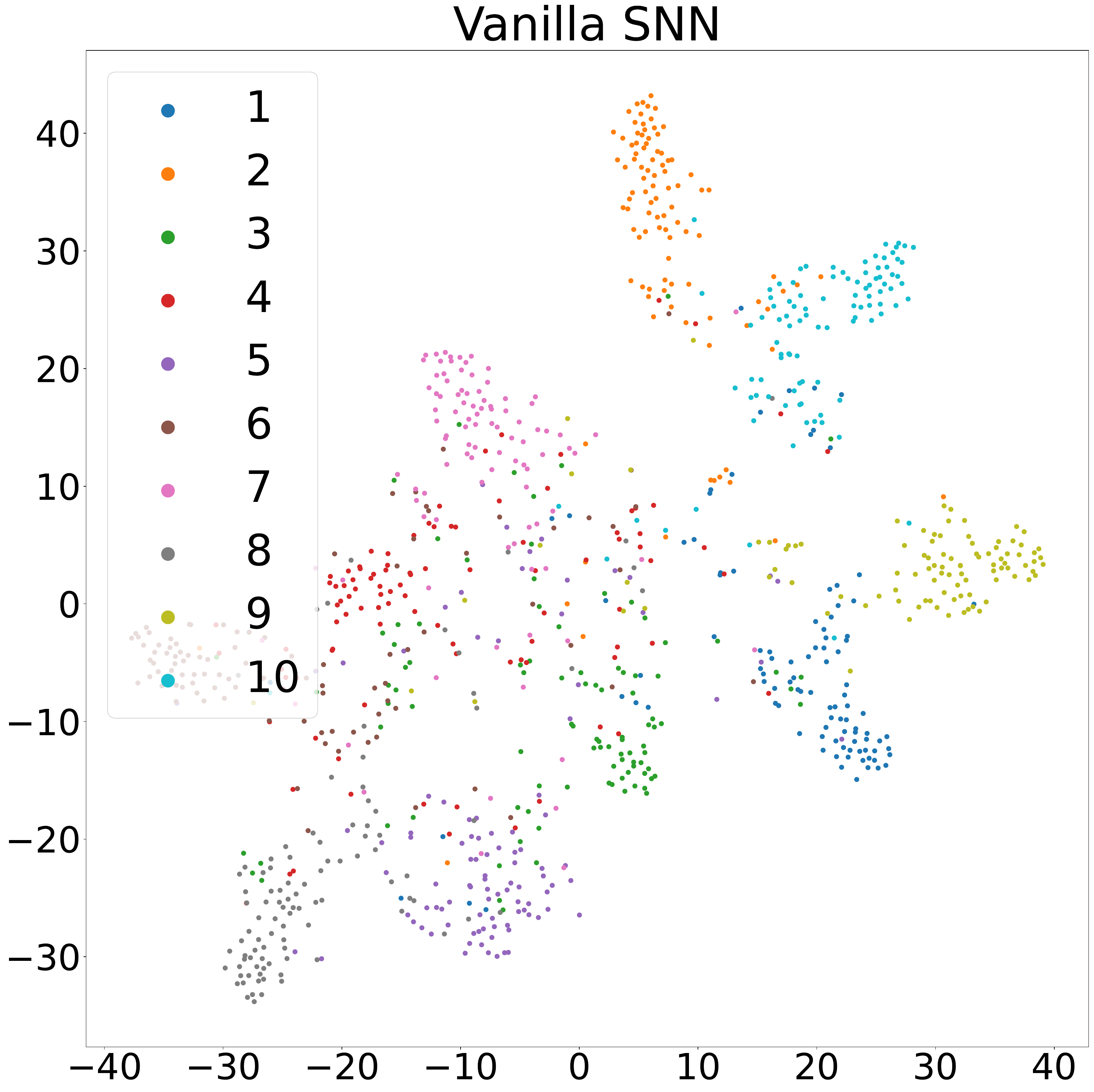}
\includegraphics[width=1.7in]{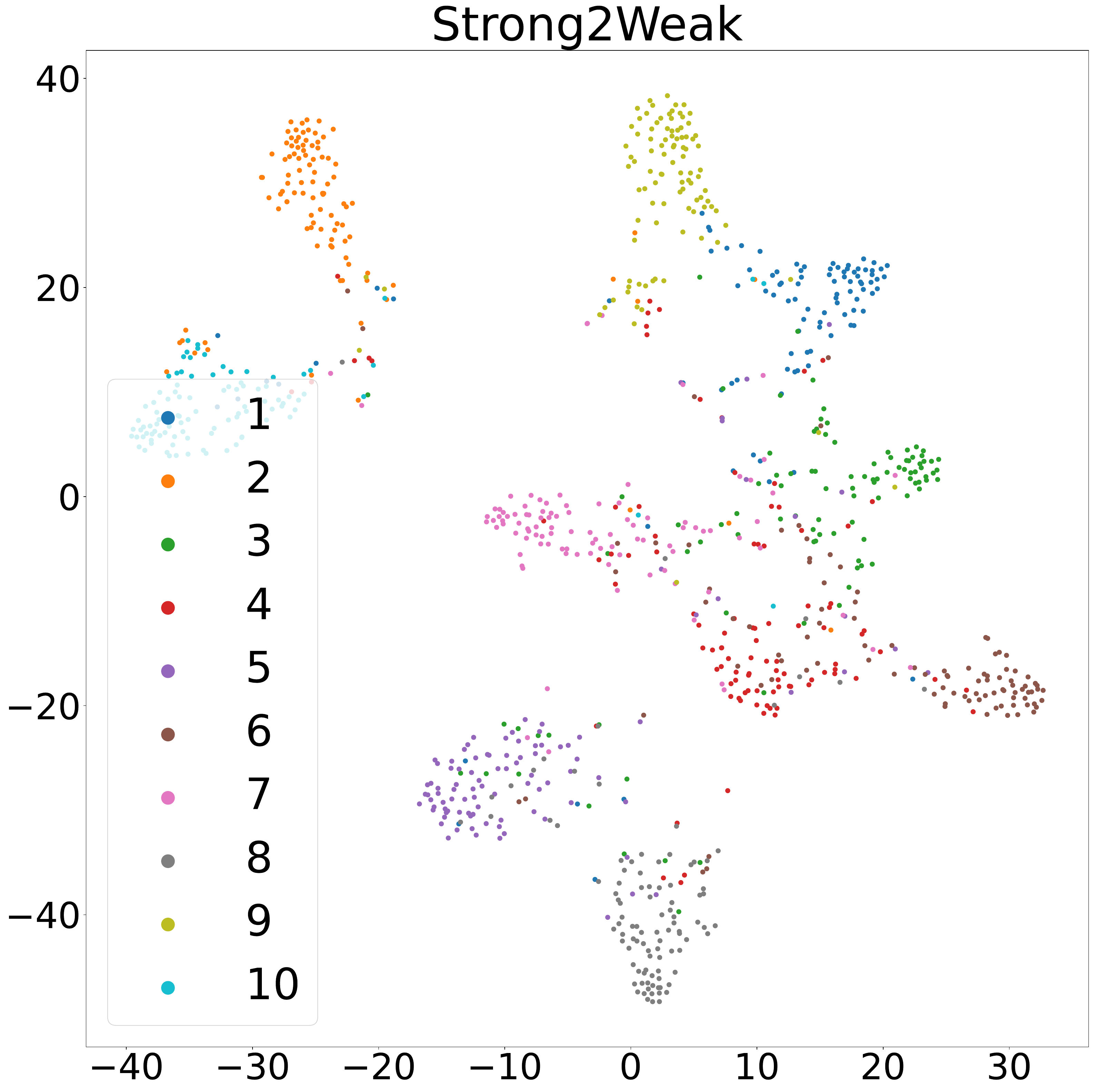}
\includegraphics[width=1.7in]{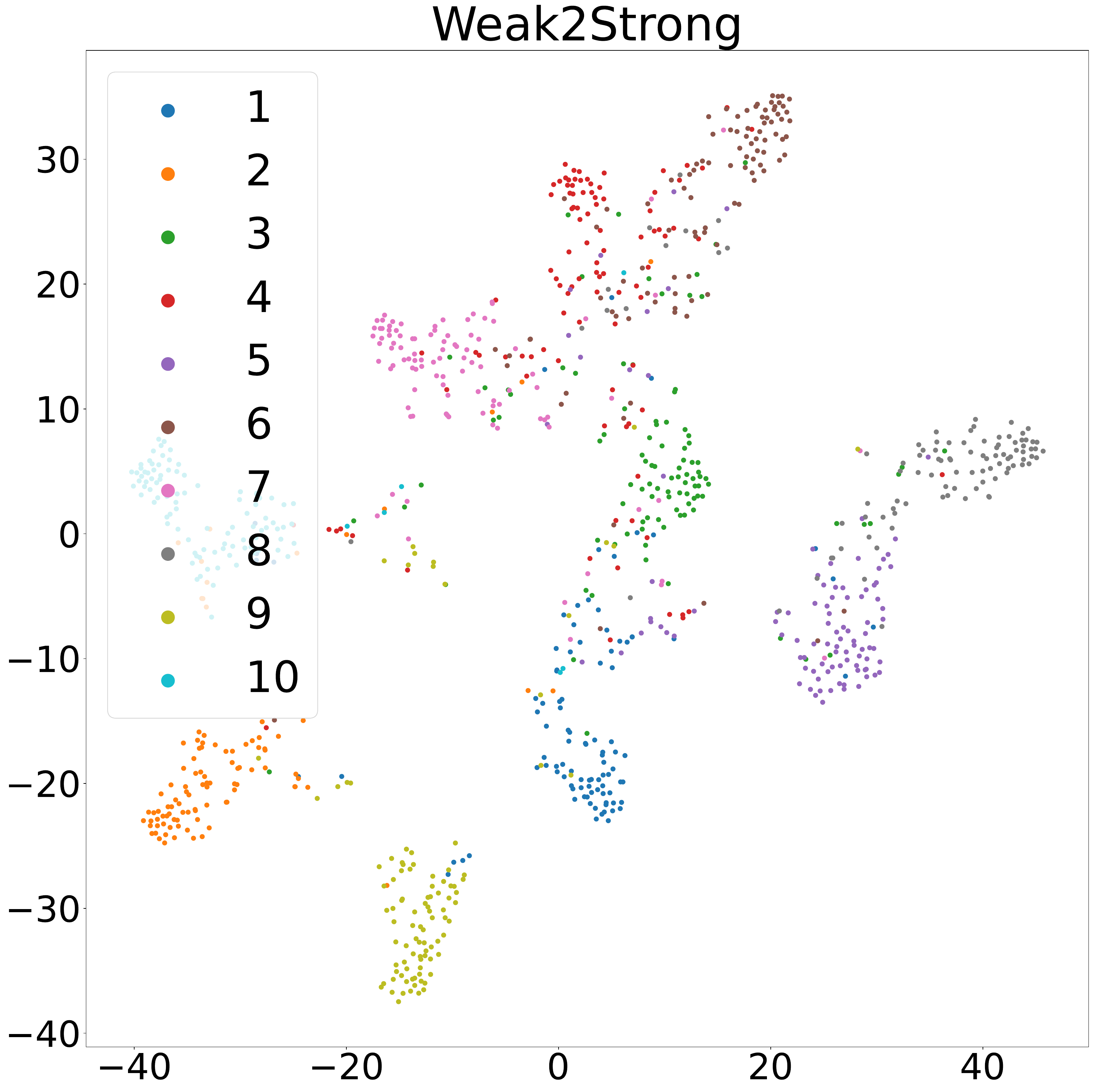}
\caption{Visualization of the overall output of the vanilla SNN, Strong2Weak distillation, and Weak2Strong distillation. Strong2Weak and Weak2Strong distillation schemes provide superior performance by allowing for more discriminable outputs than the vanilla SNN.}
\label{fig:tsne_total}
\end{figure*}

The comparison of the 2D t-SNE visualization of the overall output of the SNN is shown in Fig.~\ref{fig:tsne_total}, where our distillation schemes produce outputs that are more discriminative and thus have better performance.

\subsection{Timestep Evolution for Strong and Weak Submodels}
\begin{figure*}[h]
\centering
\includegraphics[width=5in]{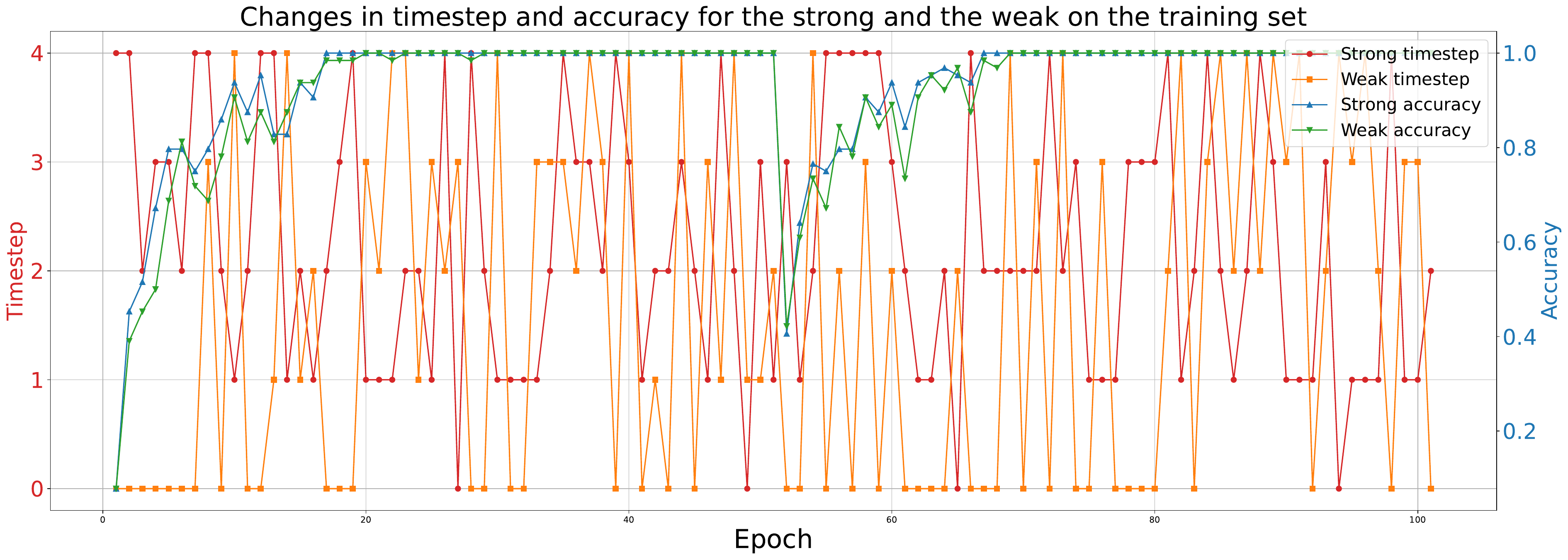}
\caption{The evolution trend of timestep indices corresponding to strong and weak submodels during training.}
\label{fig:training_timestep}
\end{figure*}

During training, we examined the timesteps corresponding to the strong and weak submodels and found that the weak submodel was not always located at the earliest timestep and that the strong model was not always the last. In fact, the weak submodel appears at every timestep, as shown in Fig.~\ref{fig:training_timestep}. However, it appears most frequently at the earliest timestep and less frequently at other timesteps. This is consistent with our experimental results: the vanilla SNN performs worst at the earliest timestep, while our method significantly improves the performance of the first timestep, so it is not always the weakest. The strong submodel is less likely to be selected in the first timestep, but is typically selected for subsequent timesteps. This demonstrates that our method of evaluating submodels based on confidence and dynamically selecting timesteps for distillation is capable of adapting to the training of SNNs to a certain extent. Otherwise, it would degrade to distillation with specified timesteps.

Additionally, we compare distillation with specified timesteps. We construct Last2First and First2Last distillation schemes under the assumption that the first timestep is the weakest and the last timestep is the strongest. The comparative results are shown in Table~\ref{com_random}. The results show that specifying these two timestep distillations is much less effective than our confidence selection distillation. This highlights the importance of dynamically selecting teacher and student timesteps.

\section{Appendix D: Additional Experiments}
\label{app_D}

\subsection{Influence of Loss Function Coefficients}
By default, the distillation function coefficients for Strong2Weak and Weak2Strong are set to 1.0. In Table~\ref{com_coe}, we explore the performance for other values of the loss function coefficients. The results show that the accuracy varies slightly with the coefficients, but remains stable overall (except for the weights of 0.1 and 2.0). The adjustment of the loss coefficients based on well-established theories should further contribute to the performance of the proposed methods, taking into account the relevant work in the field of knowledge distillation.

\begin{table}[h]
 \centering
  \caption{Influence of loss function coefficient values on performance (\%). The accuracy varies slightly with the coefficients, but remains stable overall.}
 \label{com_coe}
 \begin{tabular}{ccc|cc}
  \toprule
 \multirow{2}{*}{Coefficient} & \multicolumn{2}{c}{CIFAR10-DVS} & \multicolumn{2}{c}{DVS-Gesture}\\ & S2W & W2S & S2W & W2S\\
  \midrule
  0.1 & 77.27 & 76.43 & 93.06 & 91.67\\
  0.5 & 78.30 & 78.57 & 92.13 & 91.55\\
  1.0 & 78.93 & 79.33 & 91.43 & 91.20\\
  1.5 & 79.13 & 78.53 & 90.51 & 90.16\\
  2.0 & 79.17 & 78.17 & 89.24 & 87.96\\
  \bottomrule
 \end{tabular}
\end{table}

\subsection{Influence of Distillation Temperature}

The default distillation temperature is set to 2.0. Table~\ref{com_temperature} shows the influence of distillation temperature on performance. The results indicate that adjusting the distillation temperature within a reasonable range only causes slight fluctuations in performance without leading to significant degradation. These results also demonstrate that our method can achieve satisfactory performance without the need for deliberate adjustments to the distillation temperature. However, when the temperature or distillation weight falls outside the normal range (e.g., a temperature of 0.5), the performance of our method deteriorates. Nevertheless, it consistently outperforms vanilla SNNs.

\begin{table}[h]
 \centering
  \caption{Influence of distillation temperature on performance (\%).}
 \label{com_temperature}
 \begin{tabular}{ccc|cc}
  \toprule
 \multirow{2}{*}{Temperature} & \multicolumn{2}{c}{CIFAR10-DVS} & \multicolumn{2}{c}{DVS-Gesture}\\ & S2W & W2S & S2W & W2S\\
  \midrule
  0.5 & 75.30 & 76.77 & 91.09 & 91.32\\
  1.0 & 77.63 & 78.43 & 92.13 & 91.44\\
  2.0 & 78.93 & 79.33 & 91.43 & 91.20\\
  3.0 & 78.27 & 78.43 & 90.86 & 91.21\\
  5.0 & 78.75 & 79.03 & 92.01 & 91.67\\
  \bottomrule
 \end{tabular}
\end{table}

\subsection{Comparison with High Accuracy Submodel Teacher}

After deconstructing the SNN into multiple submodels, the naive distillation method is to select the submodel with the highest accuracy among them as the teacher guiding the submodel with the lowest accuracy. However, the naive method has three weaknesses that limit its availability and performance.

\begin{itemize}
    \item The distillation process is complicated by the comparison to the label on a timestep basis for each training batch.
    \item It can only be used with labeled learning, and the accuracy of each submodel cannot be evaluated if labels are missing.
    \item Relying only on accuracy to assess the strong and the weak, without being able to utilize the underlying dark knowledge, leads to limited performance. As shown in Table~\ref{com_highacc}, this naive method is inferior to our two proposed distillation schemes on both CIFAR10-DVS and DVS-Gesture.
\end{itemize}

\begin{table}[h]
 \centering
  \caption{Comparative results (\%) with high-accuracy submodel teacher (Abbreviated as HAST). This method is inferior to our two proposed distillation schemes on both CIFAR10-DVS and DVS-Gesture.}
 \label{com_highacc}
 \begin{tabular}{ccc}
  \toprule
 Method & CIFAR10-DVS & DVS-Gesture\\
  \midrule
  HAST & 78.43 & 90.62\\
  \rowcolor{S2Wcolor}Strong2Weak & 78.93 & 91.43\\
  \rowcolor{W2Scolor}Weak2Strong & 79.33 & 91.20\\
  \bottomrule
 \end{tabular}
\end{table}

\subsection{Comparison with Random Submodel Distillation}
Table~\ref{com_random} shows the comparative results between the proposed method and random submodel distillation (where two randomly selected submodels of different time steps are distilled). The experimental results show that:

(1) Random distillation outperforms the vanilla SNN, even without deliberately selecting the teacher and student. We consider this method to randomly transition between Strong2Weak and Weak2Strong, achieving an effect similar to using both simultaneously. This makes our deconstruction-after-destillation scheme more flexible and versatile. However, distillation cannot be abused entirely between submodels. As shown by the Last-to-First and First-to-Last results, performing distillation only between the first and last time steps leads to significant performance degradation.

(2) Using confidence to determine teacher and student submodels for distillation consistently achieved better performance than random distillation. This shows that although confidence is straightforward, it can be used to determine the performance of a model relatively well (though not absolutely). In fact, previous research on empirical risk minimization has shown that confidence is usually correlated with the generalization error of a model, i.e., high confidence often corresponds to low generalization error~\cite{6203595}. In contrast, random distillation cannot guarantee optimization in the direction of either strong to weak (toward low generalization error) or vice versa (toward high generalization error regularization). This easily leads to a conflict between consistency and diversity, similar to suboptimal performance when using both simultaneously.

\begin{table}[h]
 \centering
  \caption{Comparative results (\%) between the proposed method and random submodel distillation.}
 \label{com_random}
 \begin{tabular}{ccc}
  \toprule
 Method & CIFAR10-DVS & DVS-Gesture\\
  \midrule
  Vanilla SNN & 73.97 & 87.85\\
  Last-to-First & 74.75 & 89.12\\
  First-to-Last & 74.29 & 88.89\\
  Random-to-Random & 78.03 & 90.16\\
  \rowcolor{S2Wcolor}Strong2Weak & 78.93 & 91.43\\
  \rowcolor{W2Scolor}Weak2Strong & 79.33 & 91.20\\
  \bottomrule
 \end{tabular}
\end{table}

\subsection{Further Analysis of the Adversarial Robustness Gains of the Proposed Method}
Our method guides the student submodel using the output of the teacher submodel as soft labels, which can simultaneously achieve distillation and regularization effects to promote better consistency and generalization. To investigate the source of the gains in adversarial robustness, we compared our method with those of typical label smoothing and entropy regularization (with the same settings as Table~\ref{com_robust}, VGG-11 with 8 timesteps was used on CIFAR100). Further experimental results are shown in Table~\ref{com_further_robustness}. We found that using regularization methods alone can achieve decent performance on clean samples. However, pure regularization methods severely degrade and are significantly inferior to our method when faced with adversarial attacks. Therefore, we consider the source of the SNN's robustness to be the distillation that promotes internal consistency, making it immune to interference. However, it should be emphasized that regularization also plays a role to a certain extent, just as label smoothing can also achieve decent robustness.

\begin{table}[h]
 \centering
  \caption{Comparative results (\%) of the proposed method versus label smoothing and entropy regularization in terms of adversarial robustness.}
 \label{com_further_robustness}
 \begin{tabular}{ccccccc}
  \toprule
  Method  & Clean & GN & FGSM & PGD & BIM & CW\\
  \midrule
  Label smoothing & 69.92 & 68.52 & 18.29 & 8.42 & 7.89 & 23.65\\
  Entropy regularization &  70.05 & 69.11 & 17.89 & 8.17 & 7.38 & 18.88\\
  \hline
  \rowcolor{S2Wcolor}\textbf{Strong2Weak} & \textbf{70.68} & \textbf{69.53} & 21.15 & 10.30 & 9.40 & 23.40\\
  \cline{2-7}
  \rowcolor{W2Scolor}\textbf{Weak2Strong} & 70.09 & 68.87 & \textbf{21.23} & \textbf{10.70} & \textbf{9.75} & \textbf{24.08}\\
  \bottomrule
 \end{tabular}
\end{table}

\subsection{Comparative Results on Tiny-ImageNet}

The results of the proposed self-distillation schemes compared to other methods on Tiny-ImageNet are shown in Table~\ref{com_tinyimagenet}. Our method achieves competitive performance compared to other methods. In addition, our method can be seamlessly integrated with CLIF~\cite{CLIF} to further improve performance. This demonstrates the potential of our method to fuse with a wider range of neurons.

\begin{table}[!h]
 \centering
  \caption{Comparative results on Tiny-ImageNet dataset. * denotes self-implementation results with open-source code.}
 \label{com_tinyimagenet}
 \begin{threeparttable}
 \scalebox{0.94}{
 \begin{tabular}{cccc}
  \toprule
  Method  & Architecture & T & Accuracy(\%)\\
  \midrule
  Online LTL~\cite{NEURIPS2022_523caec7} & VGG-16 & 16 & 56.87\\
  ASGL~\cite{ASGL} & VGG-13 & 8 & 56.81\\
  Joint A-SNN~\cite{jointasnn} & VGG-16 & 4 & 55.39\\
  CLIF~\cite{CLIF} &  VGG-13 & 4 & 61.93\tnote{*}\\
  \hline
\multirow{2}{*}{\textbf{Strong2Weak}} & VGG-13(LIF) & 4 & 59.78\\ & VGG-13(CLIF) & 4 & \textbf{62.65}\\
  \cline{2-4}
\multirow{2}{*}{\textbf{Weak2Strong}} & VGG-13(LIF) & 4 & 59.40\\ & VGG-13(CLIF) & 4 & 62.45\\
  \bottomrule
 \end{tabular}
 }
 \end{threeparttable}
\end{table}
\end{document}